\documentclass[11pt]{article}

\usepackage{mathtools}
\usepackage{multirow}
\newcommand{\ifims}[2]{#1} 
\newcommand{\ifAMS}[2]{#1}   
\newcommand{\ifau}[3]{#3}  

\def\thetitle{Adaptive Nonparametric Clustering}
\def\theruntitle{Adaptive Nonparametric Clustering}

\def\authora{Kirill Efimov}
\def\runauthora{efimov, k.}
\def\addressa{Humboldt University Berlin,\\ IRTG 1792, Spandauer Str. 1, \\ 10178 Berlin, Germany}
\def\emaila{efimovkq@hu-berlin.de}

\def\authorb{Larisa Adamyan}
\def\runauthorb{adamyan, l.}
\def\addressb{Humboldt University Berlin, \\ IRTG 1792, Spandauer Str. 1, \\ 10178 Berlin, Germany}
\def\emailb{adamyanl@hu-berlin.de}

\def\authorc{Vladimir Spokoiny}
\def\runauthorc{spokoiny, v.}
\def\addressc{Weierstrass Institute Berlin, Mohrenstr. 39, 10117 Berlin, Germany \\
IITP RAS, HSE, Skoltech Moscow}
\def\emailc{spokoiny@wias-berlin.de}
\def\thanksc
{The research is 
supported by the Russian Science Foundation (project no. 14 50 00150).
} 

\def\theabstract{This paper presents a new approach to non-parametric cluster analysis called 
Adaptive Weights Clustering (AWC).
The idea is to identify the clustering structure by checking at different points and for different scales on departure from 
local homogeneity. 
The proposed procedure describes the clustering structure in terms of weights \( w_{ij} \)
each of them measures the degree of local inhomogeneity for two neighbor local clusters
using statistical tests of ``no gap'' between them. 
The procedure starts from very local scale, then the parameter of locality grows by some factor at each step.
The method is fully adaptive and does not require to specify the number of clusters or their structure. 
The clustering results are not sensitive to noise and outliers, 
the procedure is able to recover different clusters with sharp edges or manifold structure. 
The method is scalable and computationally feasible. 
An intensive numerical study shows a state-of-the-art performance of the method in various artificial examples and applications to text data.
Our theoretical study states optimal sensitivity of AWC to
local inhomogeneity.
}

\def\kwdp{62H30}
\def\kwds{62G10}

\def\thekeywords{adaptive weights, clustering, gap coefficient, manifold clustering}

\renewcommand{\(}{$\,}
\renewcommand{\)}{\,$}

\def\eqdef{\stackrel{\operatorname{def}}{=}}

\newcommand{\cc}[1]{\mathscr{#1}}
\newcommand{\bb}[1]{\boldsymbol{#1}}

\renewcommand{\hat}[1]{\widehat{#1}}
\renewcommand{\tilde}[1]{\widetilde{#1}}

\renewcommand{\Gamma}{\varGamma}
\renewcommand{\Pi}{\varPi}
\renewcommand{\Sigma}{\varSigma}
\renewcommand{\Delta}{\varDelta}
\renewcommand{\Lambda}{\varLambda}
\renewcommand{\Psi}{\varPsi}
\renewcommand{\Phi}{\varPhi}
\renewcommand{\Theta}{\varTheta}
\renewcommand{\Omega}{\varOmega}
\renewcommand{\Xi}{\varXi}
\renewcommand{\Upsilon}{\varUpsilon}

\def\Xv{\bb{X}}

\newcommand{\mygraphics}[3]{\begin{center}
    \resizebox{#1\textwidth}{#2\textheight}{\includegraphics{#3}}
    \end{center}
}

\usepackage{color}

\definecolor{blue(pigment)}{rgb}{0.2, 0.2, 0.6}
\definecolor{ultramarine}{rgb}{0.07, 0.04, 0.56}
\definecolor{darkspringgreen}{rgb}{0.09, 0.45, 0.27}
\definecolor{hookersgreen}{rgb}{0.0, 0.44, 0.0}
\definecolor{plum(traditional)}{rgb}{0.56, 0.27, 0.52}
\definecolor{purple(html/css)}{rgb}{0.5, 0.0, 0.5}
\definecolor{magenta(dye)}{rgb}{0.79, 0.08, 0.48}

\def\I{I\!\!I}
\def\R{I\!\!R}
\def\E{I\!\!E}
\def\N{\mathbb{N}}
\def\P{I\!\!P}

\def\kappa{\varkappa}

\def\ND{\mathcal{N}}

\def\Volume{\operatorname{Vol}}

\def\CONST{\mathtt{C} \hspace{0.1em}}
\def\cond{\, \big| \,}

\def\nsize{{n}}

\def\sumi{\sum_{i=1}^{\nsize}}

\def\ex{\mathrm{e}}

\def\Id{I\!\!\!I}
\def\Ind{\operatorname{1}\hspace{-4.3pt}\operatorname{I}}

\def\alp{\alpha}

\def\B{\cc{B}}

\def\CL{\cc{C}}
\def\dist{d}

\def\dimp{p}

\def\dimA{\mathtt{p}}

\def\dime{\dimA_{e}}

\def\dens{f}

\def\eps{\epsilon}			
\def\eps{\varepsilon}

\def\fis{\mathfrak{a}}

\def\hrate{a}
\def\hinit{h_{\mathtt{0}}}

\def\ijo{i\wedge j}
\def\ijc{i\triangle j}
\def\iju{i\vee j}
\def\ijw{ij}

\def\kmax{K}

\def\kullb{\cc{K}} 

\def\testst{T}

\def\unil{q}

\def\VsG{V_{G}}

\def\weight{w}

\def\zz{\mathfrak{z}}

\usepackage[title]{appendix}
\date{}
\usepackage{amsmath,amssymb,amsthm}
\usepackage{mathtools}
\usepackage{natbib}
\usepackage{epsfig,graphicx}
\usepackage{comment}
\usepackage{color}
\usepackage{srcltx}
\usepackage[mathscr]{eucal}
\usepackage[math]{easyeqn}
\usepackage{etoolbox}
\usepackage{hyperref}
\hypersetup{
            colorlinks,
            linkcolor=hookersgreen,
            linktoc=blue,
            citecolor=ultramarine,
            urlcolor=black,
            filecolor=black
            }
\usepackage{nameref}

\ifims{
\textheight=23cm
\textwidth=14.8cm
\topmargin=0pt
\oddsidemargin=1.0cm
\evensidemargin=1.0cm
\linespread{1.3}
\renewenvironment{abstract}
    {\centerline{\textbf{Abstract}}\bigskip
      \begin{center}
       \begin{minipage}{11cm}
        \begin{small}
    }
    {   \end{small}
       \end{minipage}
      \end{center}
     \bigskip
    }

}{ 
}

\numberwithin{equation}{section}
\numberwithin{figure}{section}
\newcounter{example}[section]
\numberwithin{example}{section}
\newcounter{remark}[section]
\numberwithin{remark}{section}
\newtheorem{theorem}{Theorem}[section]

\newtheorem{lemma}[theorem]{Lemma}

\newtheorem{exmp}[example]{Example}
\newtheorem{rmrk}[remark]{Remark}
\newenvironment{example}{\begin{exmp}\rm}{\end{exmp}}
\newenvironment{remark}{\begin{rmrk}\rm}{\end{rmrk}}

\begin{document}
\thispagestyle{empty}
\ifims{
\title{\thetitle}
\ifau{ 
  \author{
    \authora
    \ifdef{\thanksa}{\thanks{\thanksa}}{}
    \\[5.pt]
    \addressa \\
    \texttt{ \emaila}
  }
}
{  
  \author{
    \authora
    \ifdef{\thanksa}{\thanks{\thanksa}}{}
    \\[5.pt]
    \addressa \\
    \texttt{ \emaila}
    \and
    \authorb
    \ifdef{\thanksb}{\thanks{\thanksb}}{}
    \\[5.pt]
    \addressb \\
    \texttt{ \emailb}
  }
}
{   
  \author{
    \authora
    \ifdef{\thanksa}{\thanks{\thanksa}}{}
    \\[5.pt]
    \addressa \\
    \texttt{ \emaila}
    \and
    \authorb
    \ifdef{\thanksb}{\thanks{\thanksb}}{}
    \\[5.pt]
    \addressb \\
    \texttt{ \emailb}
    \and
    \authorc
    \ifdef{\thanksc}{\thanks{\thanksc}}{}
    \\[5.pt]
    \addressc \\
    \texttt{ \emailc}
  }
}

\maketitle
\pagestyle{myheadings}
\markboth
 {\hfill \textsc{ \small \theruntitle} \hfill}
 {\hfill
 \textsc{ \small
 \ifau{\runauthora}
      {\runauthora and \runauthorb}
      {\runauthora, \runauthorb, and \runauthorc}
 }
 \hfill}
\begin{abstract}
\theabstract
\end{abstract}

\ifAMS
    {\par\noindent\emph{AMS 2000 Subject Classification:} Primary \kwdp. Secondary \kwds}
    {\par\noindent\emph{JEL codes}: \kwdp}

} 
{ 
\begin{frontmatter}
\title{\thetitle}


\runtitle{\theruntitle}

\ifau{ 
\begin{aug}
    \author{\authora\ead[label=e1]{\emaila}}
    \address{\addressa \\
     \printead{e1}}
\end{aug}

 \runauthor{\runauthora}
\affiliation{\affiliationa} }
{ 
\begin{aug}
    \author{\authora\ead[label=e1]{\emaila}\thanksref{t21}}
    \and
    \author{\authorb\ead[label=e2]{\emailb}\thanksref{t22}}
    
    \address{\addressa \\
     \printead{e1}}
    \address{\addressb \\
     \printead{e2}}
    \thankstext{t21}{\thanksa}
    \thankstext{t22}{\thanksb}
    \affiliation{\affiliationa, \affiliationb} 
    \runauthor{\runauthora and \runauthorb}
\end{aug}
} 
{ 
\begin{aug}
    \author{\authora\ead[label=e1]{\emaila}\thanksref{t21}}
    \and
    \author{\authorb\ead[label=e2]{\emailb}\thanksref{t22}}
    \and
    \author{\authorc\ead[label=e3]{\emailc}\thanksref{t23}}
    
    \address{\addressa \\
     \printead{e1}}
    \address{\addressb \\
     \printead{e2}}
    \address{\addressc \\
     \printead{e3}}
    \thankstext{t21}{\thanksa}
    \thankstext{t22}{\thanksb}
    \thankstext{t23}{\thanksc}
    \affiliation{\affiliationa, \affiliationb, \affiliationc} 
    \runauthor{\runauthora, \runauthorb, and \runauthorc}
\end{aug}}

\begin{abstract}
\theabstract
\end{abstract}



\end{frontmatter}
} 

\newcommand{\tikzcircle}[2][red,fill=red]{\tikz[baseline=-0.5ex]\draw[#1,radius=#2] (0,0) circle ;}%

\graphicspath{{Figures/}}


\def\effdim{m}
\def\dimd{p}
\def\CL{\mathcal{C}}
\def\Xn{\Xv}
\def\Weight{W}
\def\supplm{Appendix}
\def\connectc{W_{\Sigma}}
\def\AWC{\textrm{AWC}}
\def\NMI{\textrm{NMI}}


\section{Introduction}

Methods for cluster analysis are well established tools in various scientific fields. 
Applications of clustering include a wide range of problems with text, multimedia, networks, and biological data. 
We refer to the book \cite{aggarwal2013data} for comprehensive overview of existing methods. 
Here we briefly overview only basic approaches in clustering, 
their advantages and problems.
First we mention the so called \textit{partitional} clustering.
These algorithms try to group points by optimizing some specific objective function, thereby using some assumptions on the data
structure.
The most known representatives of this group of methods are k-means \cite{steinhaus1956division} and its variations. 
k-means finds a local minimum in the problem of  sum of squared errors minimization.
%
The partitional algorithms generally require some parameters for initialization (number of clusters \( K \)) and also are nondeterministic by their nature. 
Also k-means usually produce spherical clusters and often fail to 
identify clusters with a complex shape. 
%
\emph{Hierarchical} methods construct a tree called dendrogram. Each level of this tree represents some partition of data with root corresponding to only one cluster containing all points. 
The base of the hierarchy consists of all singletons (clusters with only one point) which are the leaves of the tree. 
Hierarchical methods can be split into \textit{agglomerative} and \textit{divisive} clustering methods by the direction they construct the dendrogram.
The main weakness of hierarchical algorithms is irreversibility of the merge or split decisions.
%
%
%
%
\textit{Density-based} clustering was proposed to deal with arbitrary shape clusters, detect and remove noise.
It can be considered as a non-parametric method as it makes no assumptions about the number of clusters, their distribution or shapes.
DBSCAN \cite{ester1996density} is one of the most common clustering algorithms. 
DBSCAN estimates density by counting the number of points in some fixed neighborhood and retrieves clusters by grouping  dense points.
If data contains clusters with a difference in density then it is hard or even impossible to set an appropriate density level. 
Another crucial point of this approach is that the quality of nonparametric density estimation is very poor if the data dimension is large.
 \textit{Spectral} clustering methods \cite{ng2002spectral} 
 explore the spectrum of a similarity matrix, i.e. a square matrix with elements equal to pairwise similarities of the data points.
Spectral clustering can discover nonconvex clusters. 
However, the number of clusters should be fixed in advance in a proper way,
and the method requires a significant spectral gap between clusters. 
%
\textit{Affinity Propagation} (AP) \cite{frey2007clustering} is an exemplar-based clustering algorithm. 
In the iterative process AP updates two matrices based on similarities between pairs of data points. 
AP does not require the number of clusters to be determined, whereas it has limitation: 
the tuning of its parameters is difficult due to occurrence of oscillations.
\emph{Graph-based methods} represent each data point as a node of a graph whose 
 edges reflect the proximity between points. 
For a detailed survey on graph methods we refer to \cite{Zhu2005}.

After all, the following main problems arise in clustering algorithms: unknown number of clusters, nonconvex clusters, unbalance in sizes or/and densities for different clusters, stability with changing parameters. 
For big data the complexity is also very important.

A theoretical study of the clustering problem is difficult due to lack of a 
clear and unified definition of a cluster. 
Probably the most popular way of defining a cluster is based on connected level sets
of the underlined density
\cite{wishart1969,hartigan1975} or regions of relatively high density \cite{seeger2001}.
Existing theoretical bounds require to consider regular or \( r_{0} \)-standard clusters
which allows to exclude pathological cases \cite{ester1996density,rigollet2007}.
However, this approach cannot distinguish overlapping or connected clusters with 
different topological properties. 
This paper does not aim at giving a unified definition of a global cluster.
Instead we offer a new approach which focuses on local cluster properties: a cluster 
is considered as a homogeneous set of similar points. 
Any significant departure from local homogeneity is called a ``gap'' and 
a cluster can be viewed as a collection of points without a gap.  
An obvious advantage of this approach is that it can be implemented as 
a family of tests of ``no gap'' between any neighbor local clusters. 
The idea is similar to multiscale high dimensional change point analysis 
where the test of a change point is replaced by a test of a gap; cf. 
\cite{munk2013}.
The method presented in the next section involves the ideas of agglomerative hierarchical (by changing the scale of objects from small to big), density based (by nonparametric test) and affinity propagation (by iteratively updating the weights).
The ``adaptive weights'' idea originates from \textit{propagation-separation approach}
introduced in~\cite{PoSp2006} for regression types models.
In the clustering context, this idea can be explained as follows: 
for every point \( X_{i} \), the clustering procedure 
attempts to  describe its largest possible local neighborhood in which the
data is homogeneous in a sense of spatial data separation. 
Technically, for each data point
\( X_{i} \), a local cluster \( \CL_{i} \) is described in terms of binary weights \( \weight_{ij} \) and includes only points \( X_{j} \) with  \( \weight_{ij} = 1\).
Thus, the whole clustering structure of the data can be described using the matrix of weights \( W \)
which is recovered from the data. 
Usual clustering in the form of a mapping \( \Xn \mapsto \CL \)
can be done using the resulting weights \( w_{ij} \).
The weights are computed by the sequential multiscale procedure.
The main advantage of the proposed procedure is that it is fully adaptive to the 
unknown number of clusters, structure of the clusters etc. 
It applies equally well to convex and shaped clusters of different type and density.
We also show that the procedure does not produce artificial clusters (propagation effect) and 
ensures nearly optimal separation of closely located clusters with a hole of a lower density between them. 
The procedure involves only one important tuning parameter \( \lambda \), for which we suggest 
an automatic choice based on the propagation condition or on the ``sum of weights heuristic''.
Numerical results indicate state-of-the-art performance of the new method on artificial and 
real life data sets. 
The main contributions of this paper are:
\begin{enumerate}
\item 
We propose a new approach to define a clustering structure: a cluster is built by a group of samples 
with ``no gap'' inside.  
This allows to effectively recover the number of clusters and the shape of each cluster
without any preliminary information.
\item 
The method can deal with non-convex and overlapping clusters, 
it adapts automatically to manifold clustering structure and it is robust against outliers. 
\item 
The proposed procedure demonstrates state-of-the-art performance on wide range of various
artificial and real life examples 
and outperforms the popular competitive procedures even after optimising their tuning parameters. 
It is computationally feasible and the method applies even to large
datasets.
\item 
The procedure controls the probability of building an artificial cluster in a homogeneous situation. 
\item
Theoretical results claim an optimal sensitivity of the method in detecting of two or more clusters 
separated by a hole of a lower density due to multiscale nature of the procedure. 
  
\end{enumerate}

The rest of the paper is organized as follows. 
Section~\ref{SCAWC} introduces the procedure starting from some heuristics. 
Its theoretical properties are discussed in Section~\ref{SAWCprop}.
The numerical study is presented in Section~\ref{sec:eval}.
The proofs are collected in Section~\ref{SAWCproofs}. 
Some technical details 
as well as more numerical examples are postponed to \supplm.


\section{Nonparametric Clustering using Adaptive Weights}
\label{SCAWC}

Let \( \{X_{1},\ldots, X_{n}\} \subset \R^{\dimd} \) 
be an i.i.d. sample from the density \( \dens(x) \). 
Here the dimension \( \dimd \) can be very large or even infinite.
We assume for any pair \( (X_{i},X_{j}) \) that
a known distance (or non-similarity measure) \( \dist(X_{i},X_{j}) \) between \( X_{i} \) and \( X_{j} \) is given,
for instance, the Euclidean norm \( \| X_{i} - X_{j} \| \).
This is also the default distance in this paper.
The proposed procedure operates with 
the distance matrix \( \bigl( \dist(X_{i},X_{j}) \bigr)_{i,j=1}^{n} \) only.
For describing 
the clustering structure of the data, we introduce a \( n \times n  \) 
matrix of weights \( \Weight = (\weight_{ij}) \), \( i,j=1,\ldots,n \).
Usually the weights \( \weight_{ij} \) are binary and \( \weight_{ij} = 1 \) means that
\( X_{i} \) and \( X_{j} \) are in the same cluster, while \( \weight_{ij} = 0 \) indicates
that these points are in different clusters.
The matrix \( W \) is symmetric and each block of 
ones describes one cluster. 
 However, we do not require a block structure which allows to incorporate even overlapping clusters.
 For every fixed \( i \), the associated cluster \( \CL_{i} \) is given by
the collection of positive weights \( (\weight_{ij}) \) over \( j \).
One can consider a more general construction when \( \weight_{ij} \in [0,1] \)
and this value can be viewed as probability that the other point \( X_{j} \) 
is in the same cluster as \( X_{i} \).


The proposed procedure attempts to recover the weights \( \weight_{ij} \) from data, 
which explains the name ``adaptive weights clustering''. 
The procedure is sequential. 
It starts with very local clustering structure \( \CL_{i}^{(0)} \),
that is, the starting positive weights \( \weight_{ij}^{(0)} \) are limited to the closest neighbors \( X_{j} \)
of the point \( X_{i} \) in terms of the distance \( \dist(X_{i},X_{j}) \).
At each step \( k \geq 1 \), the weights \( \weight_{ij}^{(k)} \) are recomputed by means of statistical tests
of ``no gap'' between \( \CL_{i}^{(k-1)} \) and \( \CL_{j}^{(k-1)} \); see the next section.
Only the neighbor pairs \( X_{i}, X_{j} \) with \( \dist(X_{i},X_{j}) \leq h_{k} \) are checked, 
however the locality parameter \( h_{k} \) and the number of scanned neighbors \( X_{j} \) for each fixed 
point \( X_{i} \) grows in each step.
The resulting matrix of weights \( W \) is used for the final clustering.
The core element of the method is the way how the weights \( \weight_{ij}^{(k)} \) are recomputed.

\subsection{Adaptive weights \( \weight_{ij} \): test of ``no gap''}
%

Suppose that the first \( k-1 \) steps of the iterative procedure have been carried out.
This results in collection of weights \( \bigl\{ \weight_{ij}^{(k-1)}, \, {j=1,\ldots,n} \bigr\} \) 
for each point \( X_{i} \).
These weights describe a local ``cluster'' associated with \( X_{i} \).
By construction, only those weights \( \weight_{ij}^{(k-1)} \) can be positive
for which \( X_{j} \) belongs to the ball 
\( \B(X_{i},h_{k-1}) \eqdef \{ x \colon \dist(X_{i},x) \leq h_{k-1} \} \), or, equivalently,
\( \dist(X_{i},X_{j}) \leq h_{k-1} \).
At the next step \( k \) we pick up a larger radius \( h_{k} \) and 
recompute the weights \( \weight_{ij}^{(k)} \) using the previous results. 
Again, only points with \( \dist(X_{i},X_{j}) \leq h_{k} \) have to be 
screened at step \( k \).
The basic idea behind the definition of \( \weight_{ij}^{(k)} \) is to check for each pair \( i,j \) with \( \dist(X_{i},X_{j}) \leq h_{k} \)
whether the related clusters are well separated or they can be aggregated 
into one homogeneous region. 
We treat the points \( X_{i} \) and \( X_{j} \) as fixed and compute the test statistic 
\( \testst_{ij}^{(k)} \) using the weights \( w_{i\ell}^{(k-1)} \) and \( w_{j\ell}^{(k-1)} \) from the preceding step.
The test compares the data density in the union and overlap of two
clusters for points \( X_{i} \) and \( X_{j} \).
The formal definition involves the weighted empirical mass of the overlap and 
the weighted empirical mass of the union of two balls 
\( \B(X_{i}, h_{k-1}) \) and \( \B(X_{j}, h_{k-1}) \) shown on Figure \ref{fig:itestofnogap}.
\emph{The empirical mass of the overlap} \( N_{\ijo}^{(k)} \) can be naturally defined as
\begin{EQA}
	N_{\ijo}^{(k)}
	& \eqdef &
	\sum_{l\ne i,j} \weight_{il}^{(k-1)} \weight_{jl}^{(k-1)} .
\label{Nijodef}
\end{EQA}
In the considered case of indicator weights \( \weight_{ij}^{(k-1)} \), this value is indeed equal to
the number of points in the overlap of \( \B(X_{i}, h_{k-1}) \) and \( \B(X_{j}, h_{k-1}) \)
except \( X_{i},X_{j} \).
Similarly, 
the \emph{mass of the complement} is defined as 
\begin{EQA}
\label{Nijcompldef}
	N_{\ijc}^{(k)}
	& \eqdef &
	\sum_{l\ne i,j} \Bigl\{ 
		\weight_{il}^{(k-1)} \Ind\bigl( X_{l} \not\in \B(X_{j}, h_{k-1}) \bigr) 
		+ 
		\weight_{jl}^{(k-1)} \Ind\bigl( X_{l} \not\in \B(X_{i}, h_{k-1}) \bigr) 
	\Bigr\} .
\end{EQA}
\begin{figure}
\begin{center}	
\includegraphics[height=0.12\textwidth]{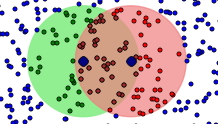}
	\quad	
\includegraphics[height=0.12\textwidth]{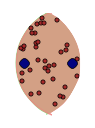}
	\quad	
\includegraphics[height=0.12\textwidth]{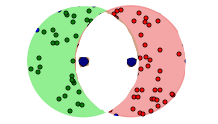}
	\quad	
\includegraphics[height=0.12\textwidth]{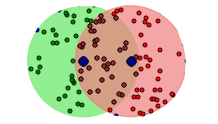}
\\[10pt]
\includegraphics[height=0.1\textheight]{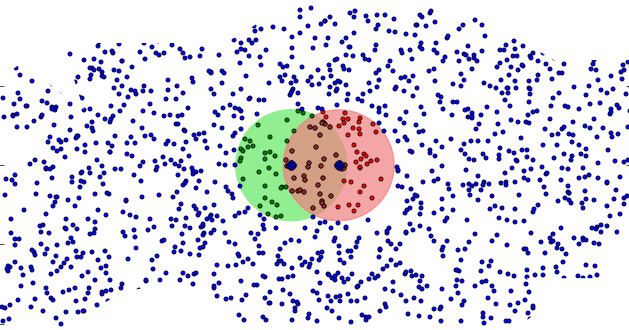}
	\quad	
\includegraphics[height=0.1\textheight]{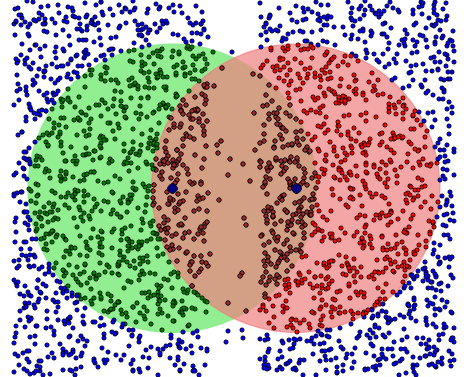}
  \includegraphics[height=0.1\textheight]{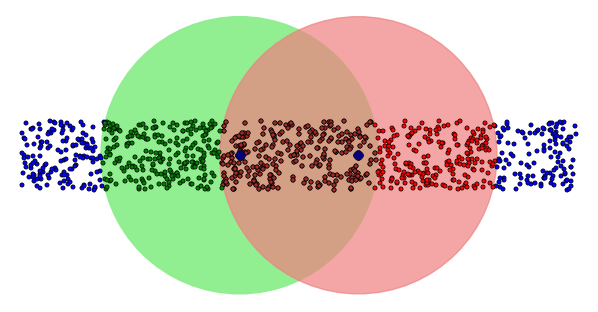}
\end{center}
\caption{Test of ``no gap between local clusters''. 
\emph{Top, from left}: Homogeneous case; \( N_{\ijo}^{(k)}; N_{\ijc}^{(k)};N_{\iju}^{(k)} \);
\emph{Bottom: from left}: Homogeneous case. ``Gap'' case. Manifold clustering.
}
\label{fig:itestofnogap}
  	\label{fig:sep5}
\label{fitestofnogap}
\end{figure}
\noindent
Note that \( N_{\ijc}^{(k)} \) is the number of points in \( \CL_{i}^{(k-1)} \)
and \( \CL_{j}^{(k-1)} \)
which do not belong to the overlap \( \B(X_{i}, h_{k-1}) \cap \B(X_{j}, h_{k-1}) \).
Finally, \emph{mass of the union} \( N_{\iju}^{(k)} \) can be defined as the sum of the mass 
of overlap and the mass of the complement:
\begin{EQA}
	N_{\iju}^{(k)}
	& \eqdef &
	N_{\ijo}^{(k)} + N_{\ijc}^{(k)} \, .
\label{Nijevr}
\end{EQA}
%
%
To measure the gap we consider the ratio of these two masses:
\begin{EQA}
	\tilde{\theta}_{\ijw}^{(k)}
	&=&
	{N_{\ijo}^{(k)}} \, \big/ \, {N_{\iju}^{(k)}} \, .
\label{thetaijkdefNN}
\end{EQA}
This value can be viewed as an estimate of the \emph{gap coefficient} \( \theta_{\ijw}^{(k)} \) which 
measures the ratio of the population mass of the overlap of two local regions 
\( \CL_{i}^{(k-1)} \) and \( \CL_{j}^{(k-1)} \) relative to the mass in their union:
\begin{EQA}
	\theta_{\ijw}^{(k)}
	&\eqdef &
	\frac{\int_{\B(X_{i},h_{k}) \, \cap \, \B(X_{j},h_{k}) } \dens(u) du}
		 {\int_{\B(X_{i},h_{k}) \, \cup \, \B(X_{j},h_{k}) } \dens(u) du} \, .
\label{thetaijkBXiBXj}
\end{EQA}
Under local homogeneity one can suppose that the density 
in the union of two balls is nearly constant.
In this case, the value \( {\theta}_{\ijw}^{(k)} \) should be close 
to the ratio \( \unil_{ij}^{(k)} \) of the volume of overlap and the volume of union of these balls:
\begin{EQA}
	\unil_{ij}^{(k)}
	& \eqdef &
	\frac
	{\Volume_{\cap}(\dist_{ij}, h_{k-1})}
	{2\Volume(h_{k-1}) - \Volume_{\cap}(\dist_{ij}, h_{k-1})}, \quad
\label{Tijkqijk}
\end{EQA}	
where \( \Volume(h) \) is the volume of a ball with radius \( h \) 
and \( \Volume_{\cap}(\dist, h) \) is the volume of the intersection of two balls
with radius \( h \) and distance \( \dist_{ij} = \dist(X_{i},X_{j}) \) between centers.
The new value \( \weight_{ij}^{(k)} \) can be viewed as a randomized test of 
the null hypothesis \( H_{ij} \) of no gap between \( X_{i} \) and \( X_{j} \)
against the alternative of a significant gap. 
The gap is significant if \( \tilde{\theta}_{\ijw}^{(k)} \) is 
significantly smaller than \( \unil_{ij}^{(k)} \).
The construction is illustrated by Figure~\ref{fitestofnogap} for the homogeneous
situation and for a situation with a gap.

To quantify the notion of significance, we consider the statistical likelihood ratio test of ``no gap'' between two local clusters.
The corresponding test statistic can be motivated by the following statistical problem.
Let \( X_{1},\ldots,X_{n} \in \R^{\dimd} \) be an i.i.d. sample and \( B, C \) be two 
non-overlapping measurable sets in \( \R^{\dimd} \). 
Suppose we are interested to check the relation \( \P(B) \geq \unil\bigl\{ \P(B) + \P(C) \bigr\} \) for 
a given value \( \unil \in (0, 1) \) against the one-sided alternative 
\( \P(B) < \unil\bigl\{ \P(B) + \P(C) \bigr\} \).
Let 
\begin{EQA}
	S_{B}
	&\eqdef & 
	\sumi \Ind(X_{i} \in B),
	\quad
	S_{C}
	\eqdef  
	\sumi \Ind(X_{i} \in C).
\label{SAIndXiA}
\end{EQA}
Lemma \ref{lemma:testgap} from \supplm~\ref{SAppndA} shows that corresponding 
likelihood ratio test statistics can be written as
\begin{EQA}
	T 
	&=&
	(S_{B} + S_{C}) \, \kullb\bigl( \tilde{\theta},\unil \bigr)
	\, \bigl\{\Ind( \tilde{\theta} \leq \unil) - \Ind( \tilde{\theta} > \unil) \bigr\},
\label{TSBSCkullbttu}
\end{EQA}
where \(  \tilde{\theta} \eqdef S_{B} / ({S_{B} + S_{C}}) \), and 
\( \kullb(\theta,\eta) \) 
is the Kullback-Leibler (KL) divergence between 
two Bernoulli laws with parameters \( \theta \) and \( \eta \):
\begin{EQA}
	\kullb(\theta,\eta)
	& \eqdef &
	\theta \log \frac{\theta}{\eta} + (1 - \theta) \log \frac{1 - \theta}{1 - \eta} 
	\,\,
	.
\label{kullbthetaetadefu}
\end{EQA}
It is worth noting that the test statistic \( T \) only depends on the local sums 
\( S_{B} \) and \( S_{C} \).
One can also use the symmetrized version of the KL divergence: 
\begin{EQA}
	\kullb_{s}(\theta,\eta)
	& \eqdef &
	\frac{1}{2} \bigl\{ \kullb(\theta,\eta) + \kullb(\eta,\theta) \bigr\}
	=
	(\theta - \eta) \log \frac{\theta (1 - \eta)}{(1 - \theta) \eta} \, .
\label{kullbthetaetadefus}
\end{EQA}
Now we apply this construction to the situation with two local clusters.
The set \( B \) is the overlap of the balls \( \B(X_{i}, h_{k}) \) and \( \B(X_{j}, h_{k}) \),
while \( C \) stands for its complement within the union 
\( \B(X_{i}, h_{k-1}) \cup \B(X_{j}, h_{k-1}) \).
Then the weighted analog of the mass of the overlap \( S_{B} \) is given by \( N_{\ijo}^{(k)} \),
while \( S_{B} + S_{C} \) is extended to the mass of the union \( N_{\iju}^{(k)} \)
yielding the test statistic \( \testst_{ij}^{(k)} \) of the form
\begin{EQA}
	T_{ij}^{(k)}
	&=&
	N_{\iju}^{(k)} \,\, \kullb\bigl( \tilde{\theta}_{ij}^{(k)},\unil_{ij}^{(k)} \bigr)
	\, \bigl\{ \Ind( \tilde{\theta}_{ij}^{(k)} \leq \unil_{ij}^{(k)}) 
	- \Ind( \tilde{\theta}_{ij}^{(k)} > \unil_{ij}^{(k)}) \bigr\} .
\label{TijkNijkuijk}
\end{EQA}
\cite{PoSp2006} used a similar test of the hypothesis \( \theta_{ij}^{(k)} = \unil_{ij}^{(k)} \) 
for defining a homogeneous region within an image. 
In the contrary to that paper, we consider a one sided test with a composite null 
\( \theta_{ij}^{(k)} \geq \unil_{ij}^{(k)} \). 
%
The value \( \unil_{ij}^{(k)} \) from \eqref{Tijkqijk}
depends only on the ratio \( t_{ij}^{(k)} = {\dist_{ij}}/{h_{k-1}} \) with 
\( \dist_{ij} = \dist(X_{i},X_{j}) \).
If \( \dist(X_{i},X_{j}) = \| X_{i} - X_{j} \| \), then
\( \unil_{ij}^{(k)} \) can be calculated explicitly (\cite{li2011concise}): 
\( \unil_{ij}^{(k)} = \unil(t_{ij}^{(k)}) \) with 
\begin{EQA}[c]
\label{uniltdef}
	\unil(t)
	= 
	\left(
	2\frac
	{B\left(\frac{\dimd+1}{2},\frac{1}{2}\right)}
	{B\left(1-\frac{t^2}{4}, \frac{\dimd+1}{2},\frac{1}{2}\right)}
	 - 1
	 \right)^{-1},
\end{EQA}	
where \( B(a,b) \) is the beta-function, \( B(x,a,b) \) is the incomplete beta-function, 
and \( \dimd \) is the space dimension. 
The argument \( t \in [0,1) \) and the function can be tabulated.

The famous Wilks phenomenon \cite{wilks1938large} claims that the distribution of each test statistic \( \testst_{ij}^{(k)} \)
is nearly \( \chi^{2} \)-distributed under the null hypothesis.
This justifies the use of family of such tests \( \testst_{ij}^{(k)} \) properly scaled 
by a universal constant \( \lambda \) which is the only tuning parameter of the method. 
See below at the end of this section.

%

\subsection{The procedure}
This section presents a formal description of the procedure.
First we list the main ingredients of the method, 
then present the algorithm. 

\paragraph{A sequence of radii:}
First of all we need to fix a  growing sequence of radii
\( h_{1} \leq h_{2} \leq \ldots \leq h_{\kmax} \) which determines how fast the algorithm will come from considering very local structures to large-scale objects. 
Each value \( h_{k} \) can be viewed as a resolution (scale) of the method at step \( k \).
The rule has to ensure that the average number of screened neighbors for each \( X_{i} \) at step
\( k \) grows at most exponentially with \( k \geq 1 \).
This feature will be used to show the optimal sensitivity of the method.
A specific choice of this sequence is given in \supplm~\ref{SAppndhk}. 
Here we just assume that such a sequence is fixed under the following two conditions:
%
\begin{EQA}
	n(X_{i},h_{k+1})
	\leq 
	\hrate \, n(X_{i},h_{k}) ,
	&& \qquad 
	h_{k+1} \leq b \, h_{k} \, 
\label{hkicond}
\end{EQA}
where \( n(X_{i},h) \) is the number of neighbors of \( X_{i} \) in the ball of radius \( h \), and 
\( \hrate,b \) are given constants between 1 and 2.
Our default choice is \( \hrate = \sqrt{2} \), \( b = 1.95 \) which ensures a non-trivial overlap of any 
two local clusters at the step \( k \).
{Decreasing any of the parameters \( \hrate , b \) increases the number of steps \( h_{k} \) and 
thus, the computational time but can improve the separation property of the procedure. 
Our intensive numerical studies showed that the default choice works well in all examples;
no notable improvements can be achieved by tuning of these parameters.}
A geometric growth of the values \( n(X_{i},h_{k+1}) \) ensures that the total number of steps \( \kmax \)
is logarithmic in the sample size \( n \).
\paragraph{Initialization of weights:}
%
Define by \(  \hinit(X_{i}) \) the smallest 
radius \( h_{k} \) in our fixed sequence such that the number of neighbors of point \( X_{i} \) in the ball \( \B(X_{i}, \hinit(X_{i})) \) is not smaller than \( n_0 \), our default choice \( n_0 = 2 \dimd + 2 \).
Using these distances we can initialize \( \weight_{ij}^{(0)} \) as
\begin{EQA}[c]
	\weight_{ij}^{(0)} = \Ind \bigl( {\dist(X_{i},X_{j})} \leq {\max(\hinit(X_{i}),\hinit(X_{j}))}\bigr) . 
\end{EQA}

\paragraph{Updates at step \( k \):}
\label{ch:stepk}
At {step \( k \)} for \( k = 1,2,\dots, \kmax \), we update the weights 
 \( \weight_{ij}^{(k)} \) for all pairs of points \( X_{i} \) and \( X_{j} \) with distance 
 \( \dist(X_{i},X_{j}) \leq h_{k} \). 
 The last constraint allows us to recompute  only \( n \times n_{k} \) weights,
 where \( n_{k} \) is the average number of neighbors in the \( h_{k} \) neighborhood.
The weights \( \weight_{ij}^{(k)} \) at step \( k \) are computed in the form
\begin{EQA}
\label{stepkw}
	\weight_{ij}^{(k)} 
	&=& 
	\Ind \bigl( {\dist(X_{i},X_{j})} \leq {h_{k}}\bigr) \,
	\Ind \bigl( {\testst_{ij}^{(k)}} \leq {\lambda}\bigr)
\end{EQA}
for all points \( X_{i}, X_{j} \) with \( h_{k-1} \geq \hinit(X_{i}) \) and 
\( h_{k-1} \geq \hinit(X_{j}) \). 
The last constraint guarantees that the weights \( \weight_{ij}^{(k)} \) are computed by the algorithm 
only when the corresponding balls contain at least \( n_{0} \) points. 
The value \( \testst_{ij}^{(k)} \) is the test statistics from \eqref{TijkNijkuijk} 
for the ``no gap'' test for points 
\( X_{i} \) and \( X_{j} \). 
Here \( \lambda \) is a threshold coefficient and the only parameter for tuning. 

The output of the AWC is given by the matrix \( \Weight = \Weight^{(K)} \) at the final step \( K \) which defines the 
local cluster \( \CL(X_{i}) = (X_{j} \colon \weight_{ij} > 0)\) for each point \( X_{i} \). 
One can use these local structures to produce a partition of the data into non-overlapping blocks.  


\paragraph{Tuning the parameter \( \lambda \)} 
\label{Spropaga}
The parameter \( \lambda \) has an important influence on the performance of the method.
Large \( \lambda \)-values result in a conservative test of ``no gap'' which can lead to aggregation of inhomogeneous regions.  
In the contrary, small \( \lambda \) increases the sensitivity of the methods to inhomoheneity
but may lead to artificial segmentation.
This section discusses two possible approaches to fix the parameter \( \lambda \).
The first is not data-driven and  depends only on the data dimension \( \dimd \).
The second approach is based on the ``sum-of-weights'' 
heuristics and is completely data driven.

%
The \emph{propagation} approach which originates from \cite{SV2007} suggests to tune 
the parameter \( \lambda \) as the smallest value which ensures a prescribed level (e.g. 90\%) of correct clustering result in a very special case of just one simple cluster. 
Namely, we tune the parameter \( \lambda \) to ensure that 
the algorithm typically puts all points into one cluster
for the sample uniformly distributed on a unit ball. 
This is similar to the level condition in hypothesis testing when the procedure
is tested under the null hypothesis of a simple homogeneous cluster.
The first kind error corresponds to creating some artificial clusters, where the probability of such events is controlled by the choice of \( \lambda \).
The construction only guarantees the right performance of the method (\( \weight_{ij}^{(k)} \approx 1 \)) 
in the very special case of locally constant density.
However, this situation is clearly reproduced for any local neighborhood lying within a large 
homogeneous region. 
Therefore, the propagation condition yields a right performance of the procedure within each homogeneous region.

Another way of looking at the choice of \( \lambda \)
called \emph{``sum-of-weights heuristic''}, is based on the effective cluster volume 
given by the total sum of final weights \( \weight_{ij}^{(\kmax)} \) over all \( i,j \).
Let \( \weight_{ij}^{(\kmax)}(\lambda) \) be the final weights obtained by the procedure 
with the parameter \( \lambda \). 
Define
\begin{EQA}
	S(\lambda)
	& \eqdef &
	\sum_{i,j=1}^{n} \weight_{ij}^{(\kmax)}(\lambda) .
\label{SlambdadefCL}
\end{EQA}
Small \( \lambda \)-values lead to artificial clustering with many small blocks of ones and all zeros 
outside of these blocks. 
The corresponding \( S(\lambda) \) will be small as well. 
An increase of \( \lambda \) yields larger homogeneous blocks and thus, 
a larger value \( S(\lambda) \).
Such behavior is typically observed until \( \lambda \) reaches a reasonable value, then the cluster structure stabilizes and any further moderate increase of \( \lambda \) does not affect \( S(\lambda) \).
For big \( \lambda \), the procedure starts to aggregate 
two or more clusters into one, this leads to a jump in \( S(\lambda) \). 
So, a proposal is to pick up the smallest \( \lambda \)-value corresponding to a plateau in the graph of 
\( S(\lambda) \).
In the case of complex cluster structure, one can observe several plateaux, 
with the corresponding \( \lambda \)-value for each plateau.
Then we recommend to check all those \( \lambda \)-values and compare the obtained 
clustering results afterwards.
See \supplm~\ref{SAppndSG} for some numerical examples.

%

\subsection{Algorithm complexity}

The preliminary step of our algorithm requires to
fix the sequence of radii \( \{h_{k}\}_{k=0}^{\kmax} \), build the distance matrix and initialize the matrix of weights. The last is updated on each step of the algorithm.
Suppose the average number of neighbors for each \( X_{i} \) at step \( k \) is \( n_{k} \).
Then finding the first \( n_{\kmax} \) neighbors for each point costs 
\( O(n \, n_{\kmax} \, \log n) \). 
At step \( k \) we need to compute \( 0.5 n \, n_{k} \) statistics \( \testst_{ij}^{(k)} \). 
Calculation of all values \( N_{\ijo}^{(k)}, N_{\ijc}^{(k)} \, \) costs \( O(n \, n_{k}^2) \).
As a result the overall complexity of step \( k \) is \( O(n n_{k}^2) \).
Note that the local nature of the procedure allows to effectively use parallel computations.  
In our approach the radii \( h_{k} \) are fixed in a way that ensures exponential growth of \( n_{k} \) . 
It results in \( \kmax = O(\log n) \) steps and furthermore, complexity of all steps is determined by the last step: \( O(n n_{\kmax}^2) \). 
In our experiments, the sample size was not large \( n 
\leq  2000 \) and we used \( n_{\kmax} = n \). 
For large datasets, one should use \( n_{\kmax} \ll n \). 
Then at the last step we will only catch the local clustering structure for each point and then ``recover'' the global structure by extracting the connected components.

\section{AWC properties}
\label{SAWCprop}
This section discusses some important properties of the AWC method.

\subsection{Propagation for regions with a non-constant density}
The procedure is calibrated to ensure the propagation within regions with 
a constant density. 
It is important to understand how far this property can be extended for
a non-constant density.
Symmetricity arguments allow to easily extend propagation effect to the case 
of a linear density. 
In the univariate case, one can make a further step and show this property for regions 
with a concave density.

\begin{theorem}
\label{Tpropconcdens}
Let the observations \( X_{i} \) be i.i.d. in \( \R^{\dimd} \), let the data density \( \dens(x) \) be supported on a region \( V \). 
Consider two cases: 1)  \( \dimd = 1 \) and the density \( \dens(x) \) is concave;
2) \( \dimd \) is arbitrary and \( \dens(x) \) is linear.
If \( \lambda > \CONST \log n  \) for some absolute constant \( \CONST \), then 
with a probability at least \( 1 - 2/n \), it holds 
\( \weight_{ij}^{(k)} = 1 \) at any step \( k \) of the procedure.
\end{theorem}

Numerical examples illustrating this and the further results are presented in Section~\ref {sec:onegausscl}.
The proofs are collected in Section~\ref{SAWCproofs}.

\subsection{Separation with a hole}
\label{SseparAWC}
Now we discuss the ``separation'' effect between clusters
for one particularly important situation, 
when two homogeneous regions are separated by a hole with slightly smaller density
and we compute the weight \( \weight_{ij}^{(k)} \) by \eqref{stepkw} for two points from different regions each close to the hole; 
see Figure~\ref{fitestofnogap} right.
Let \( V \) be a set with the volume \( |V| \) and \( G \) be a splitting hole 
with volume \( |G| \)
such that \( \VsG \eqdef V \setminus G \) consists of two disjoint regions.
To be more specific, 
consider two uniform clusters separated by some area (hole) of a lower density. 
Let \( \dens_{G} \) denote the density on \( G \) and \( \dens_{\VsG} \) 
on the complement \( V \setminus G \).
We assume the relation \( \dens_{G} = (1 - \eps) \dens_{\VsG} \) for a small value \( \eps \).
The separation effect would mean that for any pair of points \( X_{i} \) and \( X_{j} \) from 
different clusters, the statistical test detects this situation leading to a big value of the 
test statistic \( \testst_{ij}^{(k)} \) and to a vanishing weight \( \weight_{ij}^{(k)} \).  
The next two theorems answer the following question:
what is the smallest depth parameter \( \eps \) of the hole which enables a consistent and precise separation?
First we establish a lower bound.

\begin{theorem}
\label{TmainCLseplb}
Let the data support \( V \) contain a fixed hole \( G \),
and the data density \( \dens(\cdot) \) be equal to \( \dens_{1} \) on the complement \( V \setminus G \) and to \( \dens_{G} \) on \( G \) with \( \dens_{G} = (1 - \eps) \dens_{1} \). 
Let \( \eps = \eps_{n} \) as the sample size \( n \to \infty \).
If \( n \eps_{n}^{2} \leq \CONST \) for a fixed constant \( \CONST > 0 \),
then it is impossible to consistently separate the cases with \( \eps = 0 \)
(no gap) and \( \eps = \eps_{n} \).
\end{theorem}

For an upper bound, we need a more specific description of the shape of the region \( V \)
on which the data is supported.
Namely we assume that \( V \) is composed of two regions \( V_{1} \) and \( V_{2} \) 
of higher density \( \dens_{1} \) separated by a 
hole \( G \) with a slightly smaller density \( \dens_{G} \) and the volume and the shape of all three subregions \( V_{1}, V_{2}, G \) is
nearly the same. 
The next result heavily uses the multiscale nature of the procedure. 
Namely we focus on the steps when the bandwidth \( h_{k} \) approaches the global 
bandwidth \( h_{\kmax} \).
For two points \( X_{i} \) and \( X_{j} \) from different regions, this allows to assume 
that the union of two balls \( \B(X_{i},h_{k}) \) and \( \B(X_{j},h_{k}) \) contains 
the whole domain \( V \), while their overlap contains \( G \). 
We show that for such a configuration the computed weights 
\( \weight_{ij}^{(k)} \) typically vanish provided that \( n \eps^{2} \geq \CONST \log(n) \). 

\begin{theorem}
\label{TmainCLsep}
Let a set \( V \) be split by a hole \( G \) with \( \delta = |G|/|V| \geq 1/3 \).
Let the data density \( \dens(\cdot) \) fulfill \( \dens(x) \leq \dens_{G} \) for 
\( x \in G \) and \( \dens(x) \geq \dens_{1} \) for \( x \in V \setminus G \)
with \( \dens_{G} \leq (1 - \eps) \dens_{1} \). 
Let \( X_{i} \in V_{1} \), \( X_{j} \in V_{2} \) be two sample points from different regions 
and let for some \( k \leq \kmax \) and the corresponding bandwidth \( h_{k} \), it hold 
\begin{EQ}[c]
	\B(X_{i},h_{k}) \cup \B(X_{j},h_{k}) 
	= 
	V,
	\quad
	\B(X_{i},h_{k}) \cap \B(X_{j},h_{k}) 
	 \supseteq 
	G,
	\\
	|V|/3
	 \leq 
	|\B(X_{i},h_{k}) \cap \B(X_{j},h_{k})| 
	\leq  
	|V|/2 .
\label{BXiXjhkV23}
\end{EQ}
If 
\(
n \eps^{2} 
	 \geq 
	\CONST \log(n)
\)
for a fixed sufficiently large constant \( \CONST \), 
then the AWC procedure assigns the weight \( \weight_{ij}^{(k)} = 0 \) 
with a high probability.
\end{theorem}

The conditions \eqref{BXiXjhkV23} of the theorem on the shape of the sets \( V \) and \( G \)
can be easily relaxed. In fact we only need that the volume of the union 
\( \B(X_{i},h_{k}) \cup \B(X_{j},h_{k}) \) to be of the order \( |V| \) and significantly larger 
than the volume of the overlap \( \B(X_{i},h_{k}) \cap \B(X_{j},h_{k}) \).
In its turn, this overlap has to include a massive part of the hole \( G \).
The constants 1/3 and 1/2 in the last condition can be replaced by any other two 
positive constants \( c_{1} < c_{2} < 1  \). 


\subsection{Manifold clustering and high-dimensional data}

The procedure is calibrated to ensure the propagation property which means a small probability of artificial clustering for a full dimensional homogeneous region.
It appears that this propagation property automatically extends to the case of a low dimensional manifold structure.
Suppose that the similarity measure \( \dist(X_{i},X_{j}) \) is based on the Euclidean distance between
\( X_{i} \) and \( X_{j} \).
Let also in a local vicinity of each data point 
\( X_{i} \) the remaining data concentrate in a small vicinity of a low dimensional linear subspace; see 
Figure~\ref{fig:sep5} bottom right. 
This implies that the distances \( \dist(X_{i},X_{j}) \) correspond to the effective data dimension 
\( \dime \) rather than the original dimension \( \dimd \). 
%
Here we explain why the propagation property extends to this case. 
Indeed, the test statistics 
\( \testst_{ij}^{(k)} \) are built on the base of the distance matrix \( (\dist(X_{i},X_{j})) \),
and in the manifold case, \( \testst_{ij}^{(k)} \) correspond to the effective dimension \( \dime \).
The data dimension \( \dimd \) does not enter show up there.  
There is only one place in the algorithm where the dimension \( \dimd \) appears explicitly,
namely, in the definition of the function \( \unil(\cdot) \) from \eqref{uniltdef}.
And this function decreases with \( \dimd \);
see Lemma \ref{lem:manprop} from \supplm~\ref{SAppndA}.
Artificial separation can only occur when \( \tilde{\theta}_{ij}^{(k)} < \unil_{ij}^{(k)} \).
Probability of such an event becomes very small in the case of manifold data, 
because the estimated gap coefficient \( \tilde{\theta}_{ij}^{(k)} \) corresponds to the data of
effective dimension \( \dime \), while the value \( \unil_{ij}^{(k)} \) is computed for the full 
dimension \( \dimd \).
So, one can expect that the propagation effect will be even stronger along a low dimensional manifold. 
Note however that the arguments do not apply if a low dimensional manifold crosses 
another manifold of different dimension. 
Then the procedure indicates a non-homogeneity in the same way as in the case 
of two close regions with different densities.

The manifold property allows to easily work with high-dimensional data. 
Suppose that the data dimension \( \dimd \) is large but the cluster structure corresponds to 
a low dimensional manifold of dimension \( \effdim \).
And suppose that the distance/similarity matrix \( \dist(X_{i},X_{j}) \) 
also corresponds to this manifold structure.
The definition of the adaptive weights does not rely on the dimension \( \dimd \)
except the definition of the function \( \unil(\cdot) \) from \eqref{uniltdef}.
We suggest to use the small ``effective'' dimension \( \effdim \) instead of \( \dimd \) for computing
\( \unil(t) \). 
If our guess \( \effdim \) correctly mimics the effective dimension of the data
then the AWC procedure will be properly tuned and preserve all its propagation and separation properties.
In Section \ref{sec:eval} we show the results of this approach applied on real text data.

\section{Numerical examples and evaluation}
\label{sec:eval}


This section illustrates the performance of AWC by mean of artificial and real datasets.
\subsection{Artificial data}
First examples serve to illustrate our main theoretical results.
\begin{figure}[h]
\setlength{\abovecaptionskip}{-3pt}
\hfill 
  \includegraphics[width=0.19\textwidth]{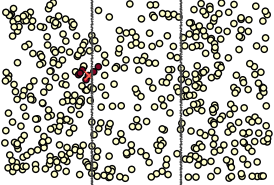}
\hfill
  \includegraphics[width=0.19\textwidth]{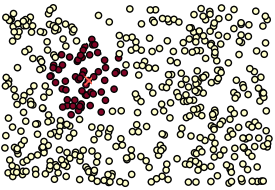}
\hfill
  \includegraphics[width=0.19\textwidth]{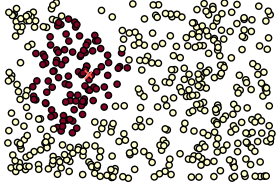}
\hfill
 \includegraphics[width=0.19\textwidth]{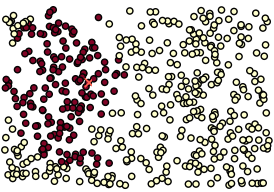}
\hfill
 \includegraphics[width=0.19\textwidth]{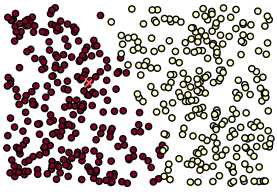}
  \caption{Steps 1, 40, 45, 47, 52. Black balls show the cluster 
  for the red point.}
  \label{fig:sequence}
\end{figure}
We start with the separation result of Theorems~\ref{TmainCLseplb} and \ref{TmainCLsep}.
Figure \ref{fig:sequence} shows the dataset composed of two uniform clusters with density \( \dens \) separated by a hole of lower density \( \dens / 2\) shown by vertical lines on the first figure.  
We fix a point on the boarder of the left cluster marked by red \( \times \).
One can see that the local cluster 
of point \( X^{*} \) at original steps spreads to the right until the radius \( r_{k} \) reaches the proper scale 
to detect the gap by our test. 
At the final step, the connections from the considered point \( X^{*} \) do not spread over the gap. As a result we have two clusters separated by a hole.
More examples on separation with a gap are presented in \supplm~\ref{SAppndSG}.

\paragraph{One Gaussian cluster}
\label{sec:onegausscl}
Suppose that the data are sampled from a standard normal law  \( \ND(0, \Id_{\dimd}) \) in \( \R^{\dimd} \).
\begin{figure}[t]
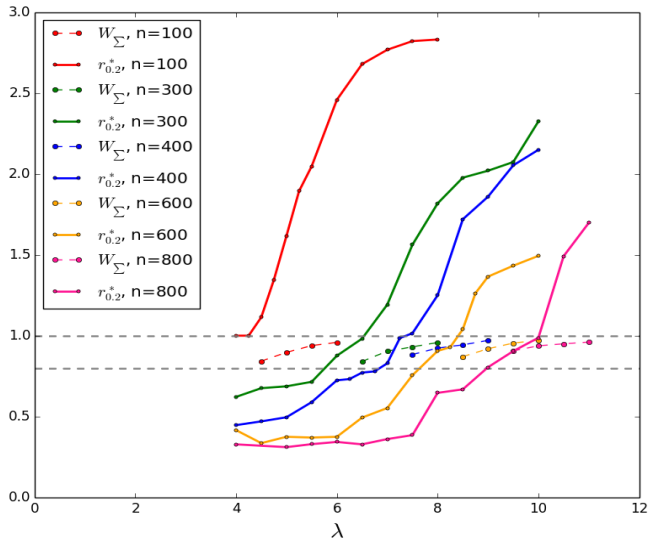

\setlength{\abovecaptionskip}{-10pt}
  \mygraphics{0.7}{0.35}
  {test/Figures_convex/alone_gaussian_1.png}
\caption{\( r^{*}_{0.2} \) for standard Gaussian data (solid lines) and averaged values of
\( \connectc(\lambda,1) \) for the uniform data on the unit disk (dashed lines) for different \( n \), \( \lambda \).}
\label{fig:onegauss}
\end{figure}
The density in this case is concave only inside a unit ball with center in 0.  
Therefore,
Theorem \ref{Tpropconcdens} implies the following behavior of AWC:
with a high probability, it detects a cluster of points associated with the unit ball.
We fix \( \dimd = 2 \).
Let \( \weight_{ij}(\lambda) \) be the final weights of the AWC procedure for a particular realization of 
the data given by AWC with parameter \( \lambda \).
Define 
the connectedness coefficient \( \connectc(\lambda,r) \) for the ball of radius \( r \):  
\begin{EQA}
	\connectc(\lambda,r) 
	&=& 
	\frac{\sum_{i,j} \weight_{ij}(\lambda) \Ind(\| X_{i} \| \leq r, \| X_{j} \| \leq r)}
		 {\sum_{i,j} \Ind(\| X_{i} \| \leq r, \| X_{j} \| \leq r)} \, .
\label{clrijwijkIXir}
\end{EQA}
Define also the radius \( r^{*} = r^{*}_{\alp} \) by the condition
\begin{EQA}
	\P\bigl( \connectc(\lambda,r^{*}_{\alp}) \geq 1-\alp \bigr)
	&=&
	1-\alp. 
\label{PWSlr080}
\end{EQA}
Solid lines in Figure~\ref{fig:onegauss} show \( r^{*}_{\alp} \) for \( \alp = 0.2 \), 
different \( \lambda \) and different sample sizes \( n = 100, 300, 400, 600, 800 \).
In addition, compute the mean of \( \connectc(\lambda,1) \) for the uniform distribution on the unit disk. 
These values are shown by the dashed lines on Figure~\ref{fig:onegauss}.
Comparing the dashed and solid lines of the same color on Figure \ref{fig:onegauss} reveals that
for a fixed \( n \), the value \( \lambda \) which guaranties 80\%-90\% connectedness in the case of uniform distribution also 
guaranties in the case of normal distribution that the radius of central cluster is close to 1. 
%
This is in complete agreement with the claim of Theorem \ref{Tpropconcdens}.

%

\paragraph{Separation for two Gaussian clusters}
Now we illustrate the 
separation properties of AWC on the example of two Gaussian clusters. 
In this case we want to check how AWC can find the possibly small gap between two clusters.
Remind that AWC is a fully nonparametric method. 
A mixture of two Gaussian distribution with nearly the same mean is still unimodal and considered as one cluster.
E.g. in the univariate case presented on Figure \ref{fig:septwogauss}, when the distance between means is less than 2 there is no gap between clusters. 
\begin{figure}[t]
\setlength{\abovecaptionskip}{-3pt}
\hfill 
  \includegraphics[width=0.22\textwidth]{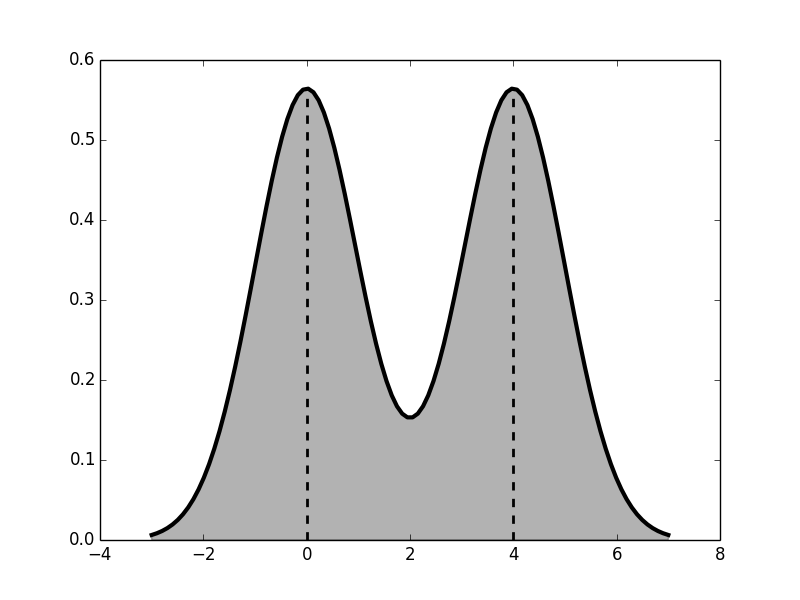}
\hfill
  \includegraphics[width=0.22\textwidth]{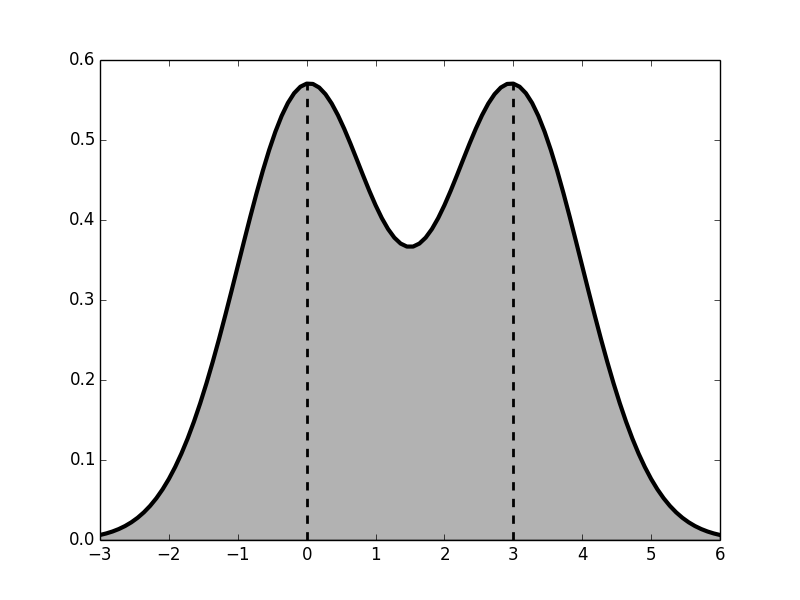}
\hfill
  \includegraphics[width=0.22\textwidth]{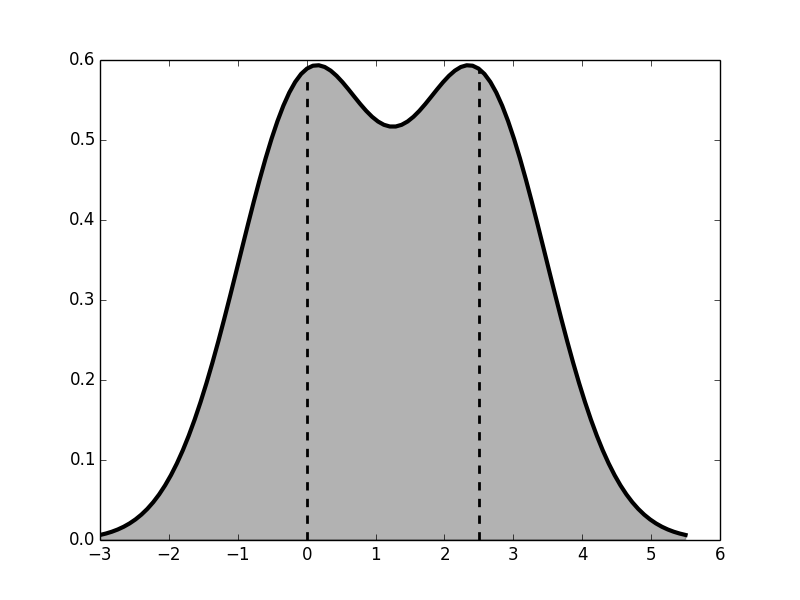}
\hfill
  \includegraphics[width=0.22\textwidth]{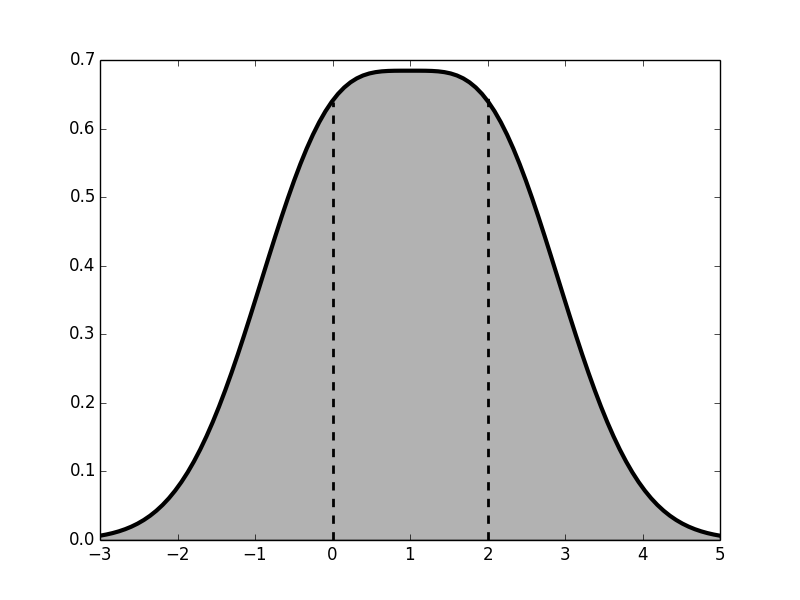}
\hfill
  \caption{Mixture of two normals with variance 1 and  distance \( D \) between means. 
  \emph{From left}:  \( D = 4; \ 3; \ 2.5;\ 2.\)}
  \label{fig:septwogauss}
\end{figure}
Let \( X_{1}, \dots, X_{n} \in \R^{2}\) be generated from standard normal distribution  \( \ND(0, \I_{2})\) and \( X_{n+1}, \allowbreak \dots, X_{2n}\) be generated from \( \ND(D, \Id_{2})\). 
Select the parameter \( \lambda \) due to suggestion of the previous section
to ensure that the radius of detected central cluster is close to 1.
Explicitly for \( n=100,200,300,400,600 \) we took \( \lambda = 4.2, 6, 6.5, 7.2, 8.3\) correspondingly.
Here we are interested in the separation error \( e_s \) from \eqref{errorsdef}.
The ideal cluster separation in this experiment is given separated by the line  \( (D / 2, y) \). 
Figure \ref{fig:twogapnorm} shows an example of such realization.
For each \( n \) and distance \( D \) we make 200 experiments. 
The averaged separation error \( e_{sp} \) as a function of distance between clusters \( D \) is shown on Figure \ref{fig:twogapnormgen}. 
One can see that the separation error remains quite high for the distance 
\( D \approx 2 \) for all considered sample sizes. 
At the same time, if the distance \( D \) exceeds 3, the procedure starts to separate 
well the Gaussian clusters without using any prior information about the structure
of the underlying density.

\begin{figure}[tb]
\centering
\setlength{\abovecaptionskip}{-3pt}
  \includegraphics[width=0.30\textwidth]{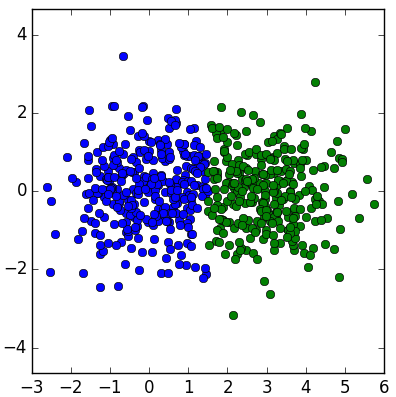}
  \qquad
  \includegraphics[width=0.30\textwidth]{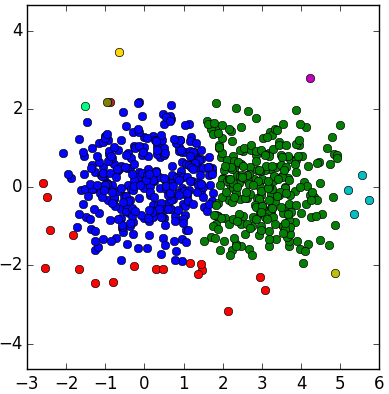}
  \caption{Mixture of two normals with \( n=300, D = 3\): ideal clustering vs AWC 
  (\( \lambda = 6.5, e_{sp} = .05 \))}
  \label{fig:twogapnorm}
\end{figure}

\begin{figure}[t]
\centering
\setlength{\abovecaptionskip}{-3pt}
  \includegraphics[width=0.60\textwidth]{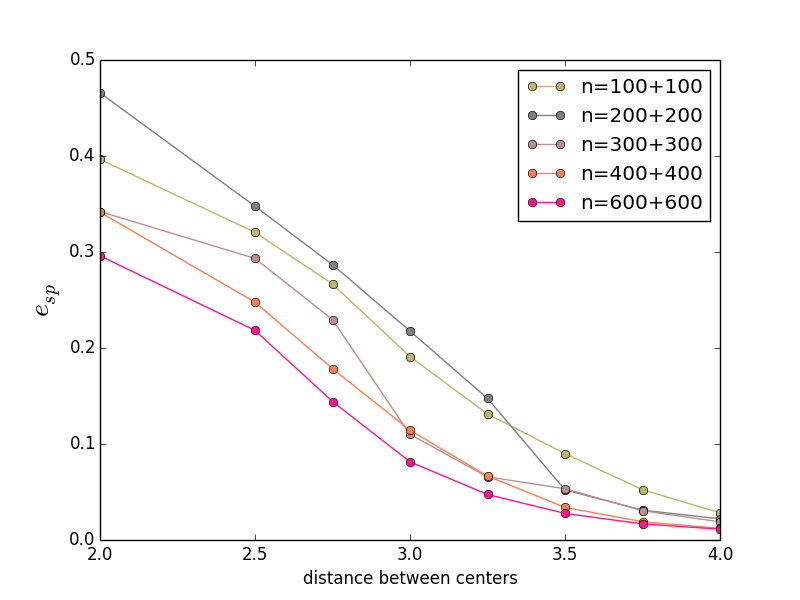}
  \caption{Separation error for the mixture of two normals}
  \label{fig:twogapnormgen}
\end{figure}

\paragraph{Performance for benchmark data}
Next we investigated the performance of AWC 
by mean of few popular artificial datasets with known cluster structure.
The tuning parameter \( \lambda \) of the AWC is selected using ``sum-of-weights'' heuristics.
First we show how AWC finds correct clusters in situations when other popular methods break down.
For comparison we used the most popular clustering software implemented in python-scikit-learn:
k-means, 
DBSCAN, spectral clustering and  affinity propagation;
\cite{pedregosa2011scikit}. 
Each method requires to fix some tuning parameter(s) and we optimized the choice for 
each particular example while the AWC is used with the automatic choice.
See \supplm~\ref{SAppndC} for details.

\begin{figure}[t]
  \includegraphics[width=0.19\textwidth,height=0.12\textheight]{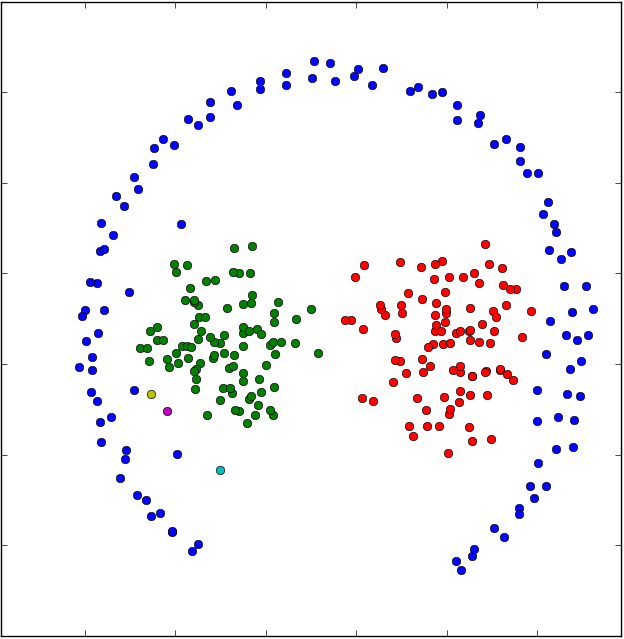}
	\hfill
  \includegraphics[width=0.19\textwidth,height=0.12\textheight]{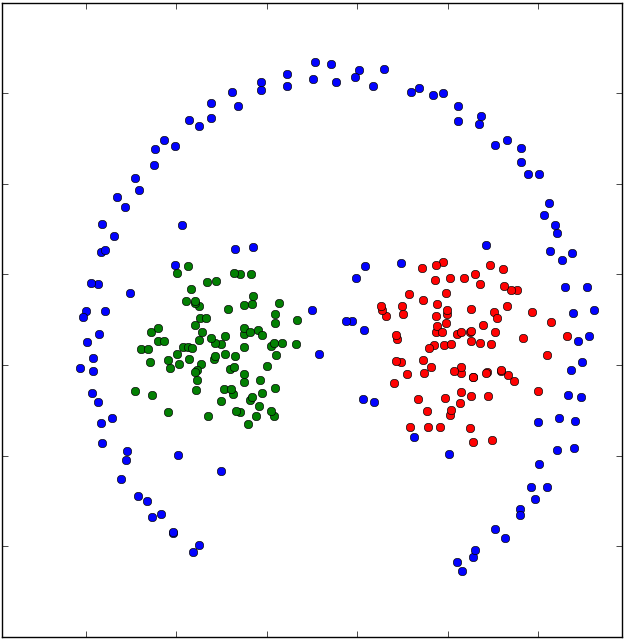}
	\hfill
  \includegraphics[width=0.19\textwidth,height=0.12\textheight]{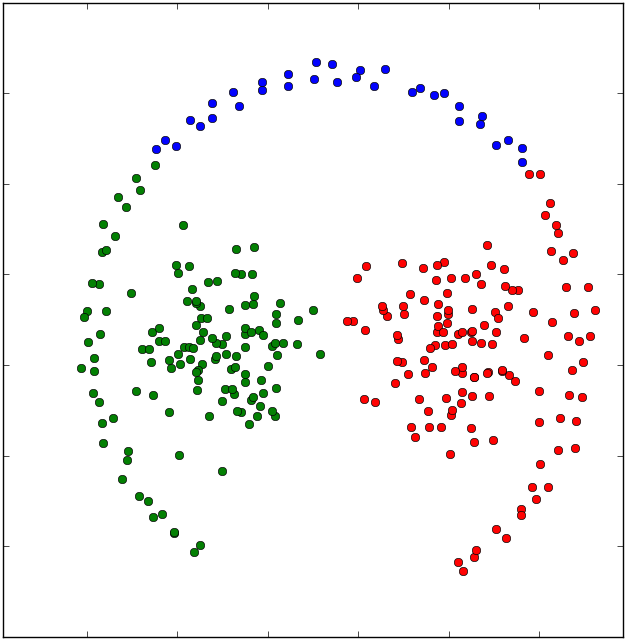}
	\hfill
  \includegraphics[width=0.19\textwidth,height=0.12\textheight]{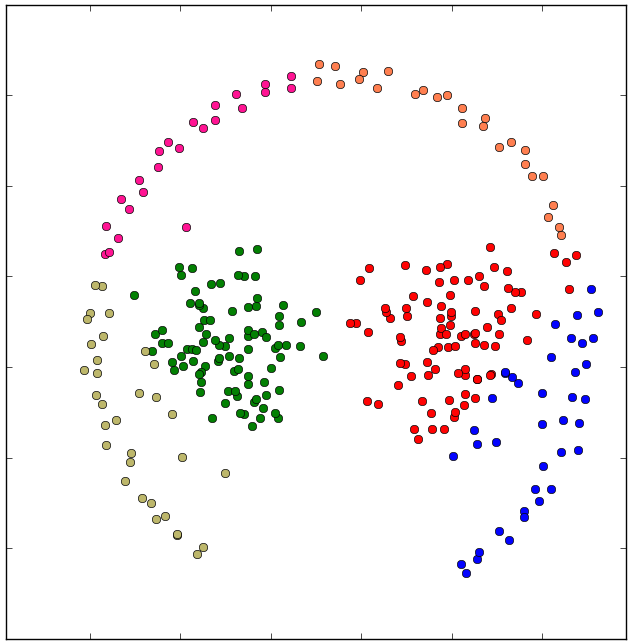}
	\hfill
  \includegraphics[width=0.19\textwidth,height=0.12\textheight]{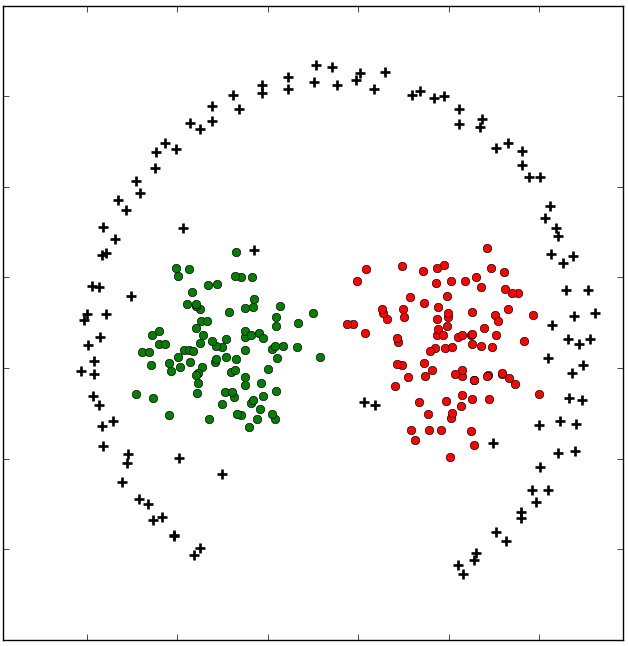}
%
%
  \\
  \includegraphics[width=0.19\textwidth,height=0.12\textheight]{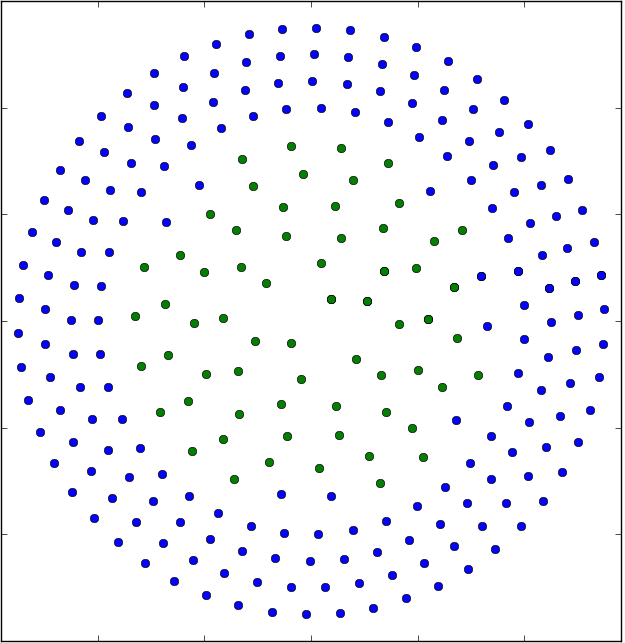}
	\hfill
  \includegraphics[width=0.19\textwidth,height=0.12\textheight]{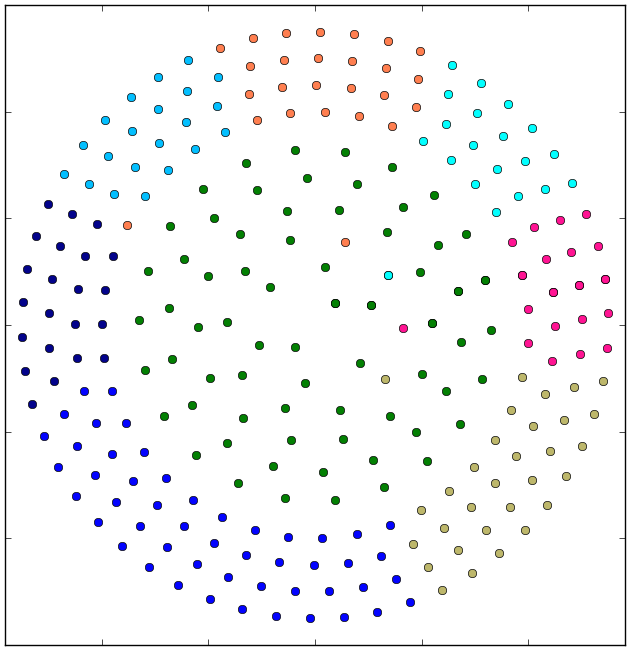}
	\hfill
  \includegraphics[width=0.19\textwidth,height=0.12\textheight]{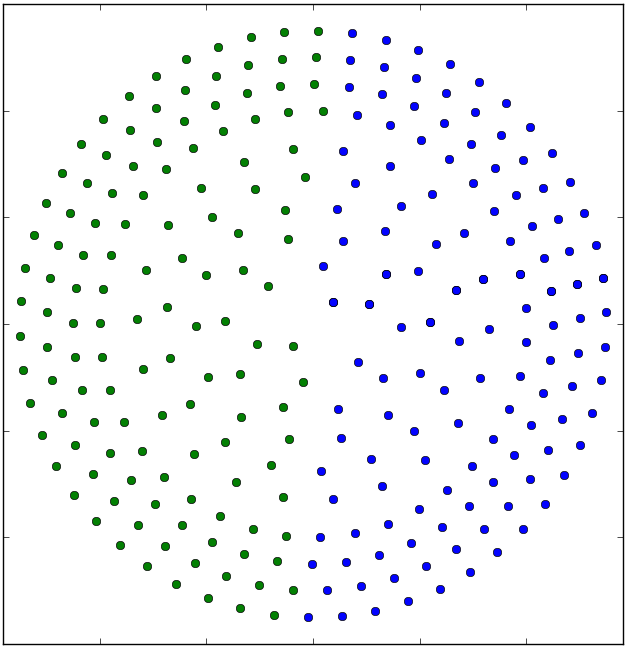}
	\hfill
  \includegraphics[width=0.19\textwidth,height=0.12\textheight]{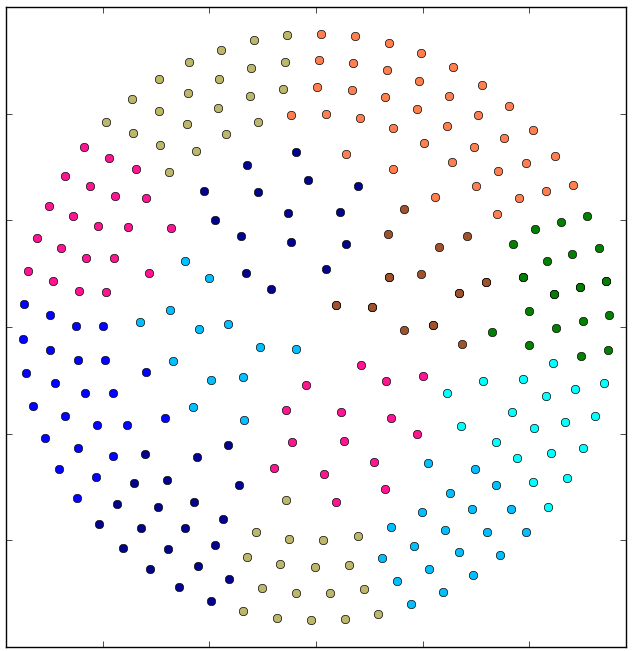}
	\hfill
  \includegraphics[width=0.19\textwidth,height=0.12\textheight]{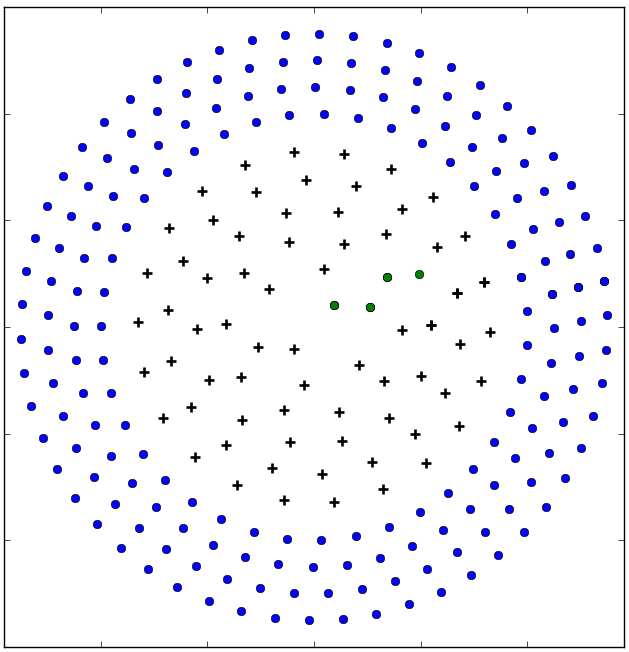}
  \caption{{Top (pathbased)}, \emph{from left}: AWC (\( \lambda=4.1 \)), Spectral (\( \sigma=0.1 \)),
  K-means (\( K=3 \)), Aff. prop. (\( D=0.5 \), \( P=-1464 \)),
  DBSCAN (\( {\eps}=2.1 \), \textit{minsp}=10); 
  {Bottom (orange)}, \emph{from left}: AWC\( (\lambda=2) \),
  spectral, K-means, affinity propagation,
  DBSCAN.}
  \label{fig:p_dbscan}
  \label{fig:o_dbscan}
\end{figure}
We consider 3 datasets.
%
The \textit{Pathbased} (300 points), Figure \ref{fig:p_dbscan} top,
consists of  two clusters with
Gaussian distribution surrounded by a  circular cluster
with an opening.  
 The \textit{Orange} dataset (268 points), Figure \ref{fig:o_dbscan} bottom, is a ball with uniform density surrounded by uniformly distributed sphere with a little bit higher density.
\textit{Compound}  \cite{zahn1971graph} is a dataset consisting of 399 points with various densities, see Figure \ref{fig:c_dbscan}. 
It contains two nearly 
 normal clusters, one small cluster surrounded 
 by a ring cluster, and a dense cluster inside big 
 sparse one.
  Figures show best performance of each comparative algorithm after parameter tuning. 
Each cluster found by the algorithms is represented by its own unique color. 
Noise points in DBSCAN result are marked by black crosses.
One can see that AWC  solves all challenges in these datasets such as 
non-convex clusters, overlapping clusters with different intensities, manifold 
clustering.
Other algorithms even after optimizing can not handle most of them
even after parameter tuning.

Other interesting examples are datasets \( DS4, DS3 \) from 
\cite{karypis1999chameleon} used for CHAMELEON hierarchical clustering algorithm. The AWC results are shown on Figure \ref{fig:chameleonawc} and we can see that AWC can 
handle these datasets as well.
Many popular within the literature artificial datasets are collected in 
\emph{https://github.com/deric/clustering-benchmark}.
AWC performance on several of them are shown on  Figure \ref{fig:manyart}. These examples include the following challenges: manifold structure (spiral data), 
the density which slowly changes inside a cluster, dense clusters with a background of low density, a dense bridge between clusters etc. 
In all examples AWC does a very good job.

%
%
\begin{figure}[t]
\begin{center}
  \includegraphics[width=0.3\textwidth,height=0.11\textheight]{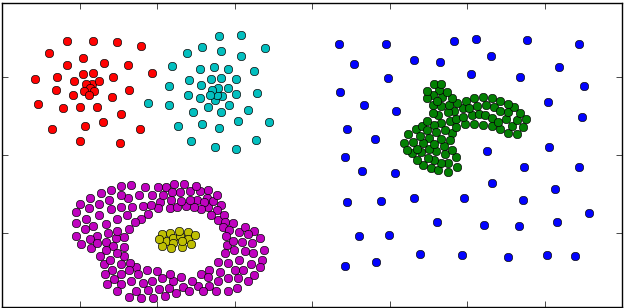}
  \includegraphics[width=0.3\textwidth,height=0.11\textheight]{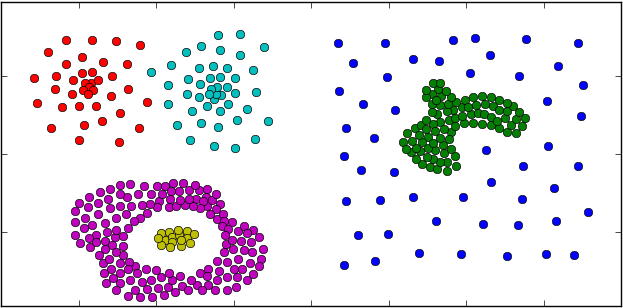}
  \includegraphics[width=0.3\textwidth,height=0.11\textheight]{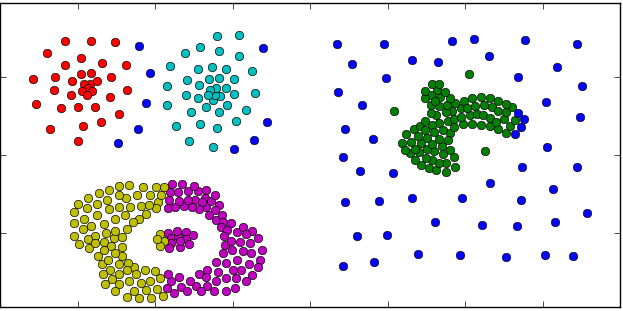}
\\
  \includegraphics[width=0.3\textwidth,height=0.11\textheight]{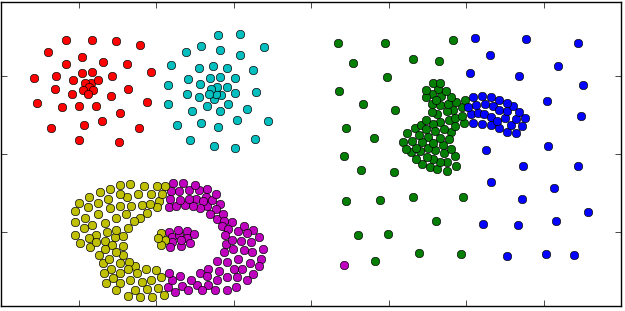}
  \includegraphics[width=0.3\textwidth,height=0.11\textheight]{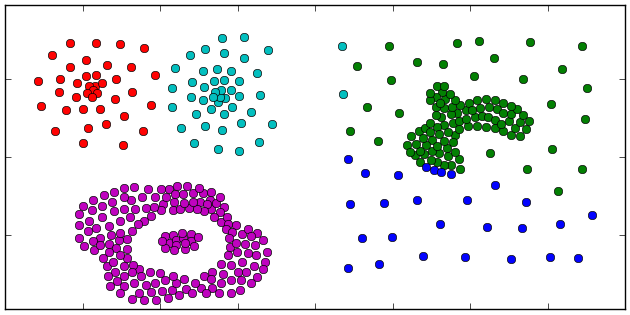}
  \includegraphics[width=0.3\textwidth,height=0.11\textheight]{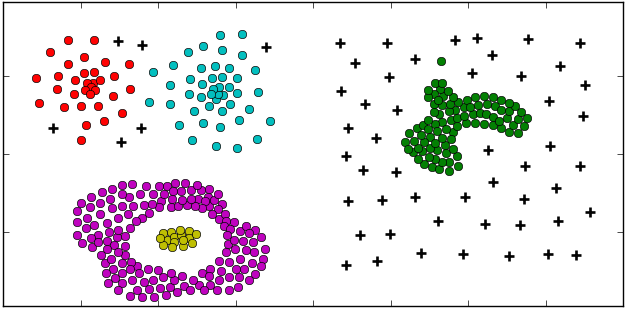}
  \caption{Compound.
  \emph{Top, from left}: Original, 
  AWC (\( \lambda=3.7 \)),
  Spectral (\(\sigma=0.1\));
  \emph{Bottom, from left}: K-means (\( K=6\)), Aff. prop. (\( D=0.5 \), \( P=-737 \)), DBSCAN (\( {\eps}=1.48 \), \textit{minsp}=3)
  }
  \label{fig:c_dbscan}
\end{center}
\end{figure}

\begin{figure}[t]
\centering
  \includegraphics[width=0.49\textwidth,height=0.19\textheight]{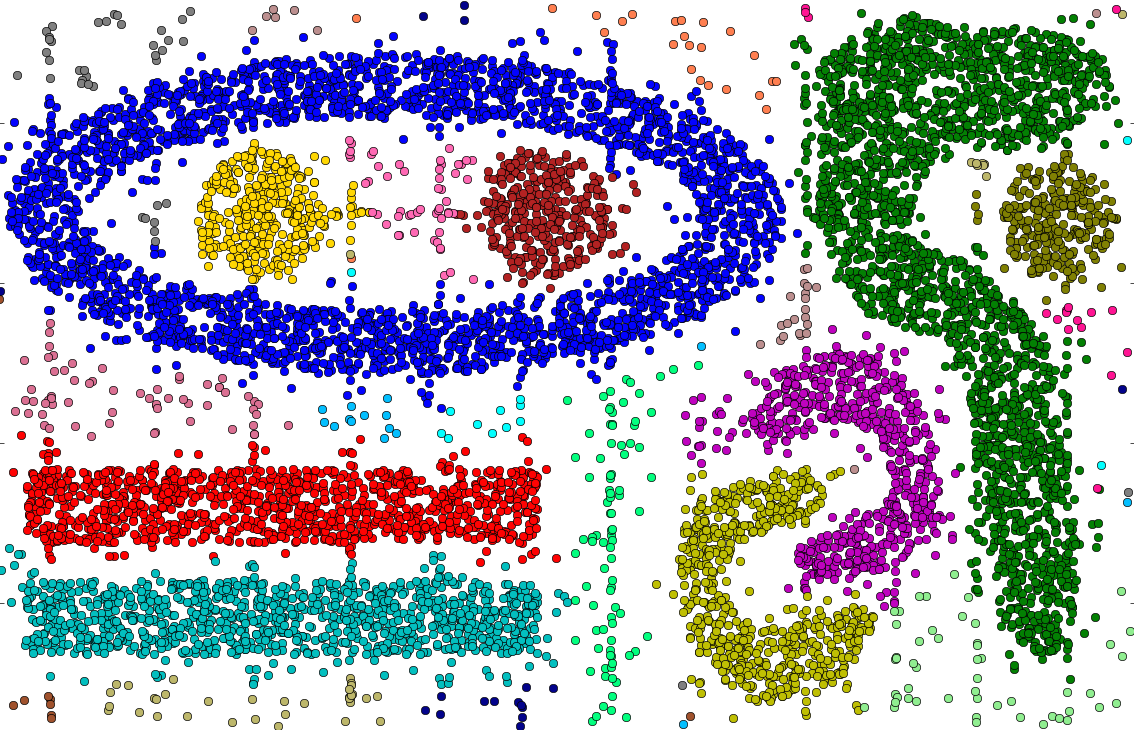}
  \hfill
  \includegraphics[width=0.49\textwidth,height=0.19\textheight]{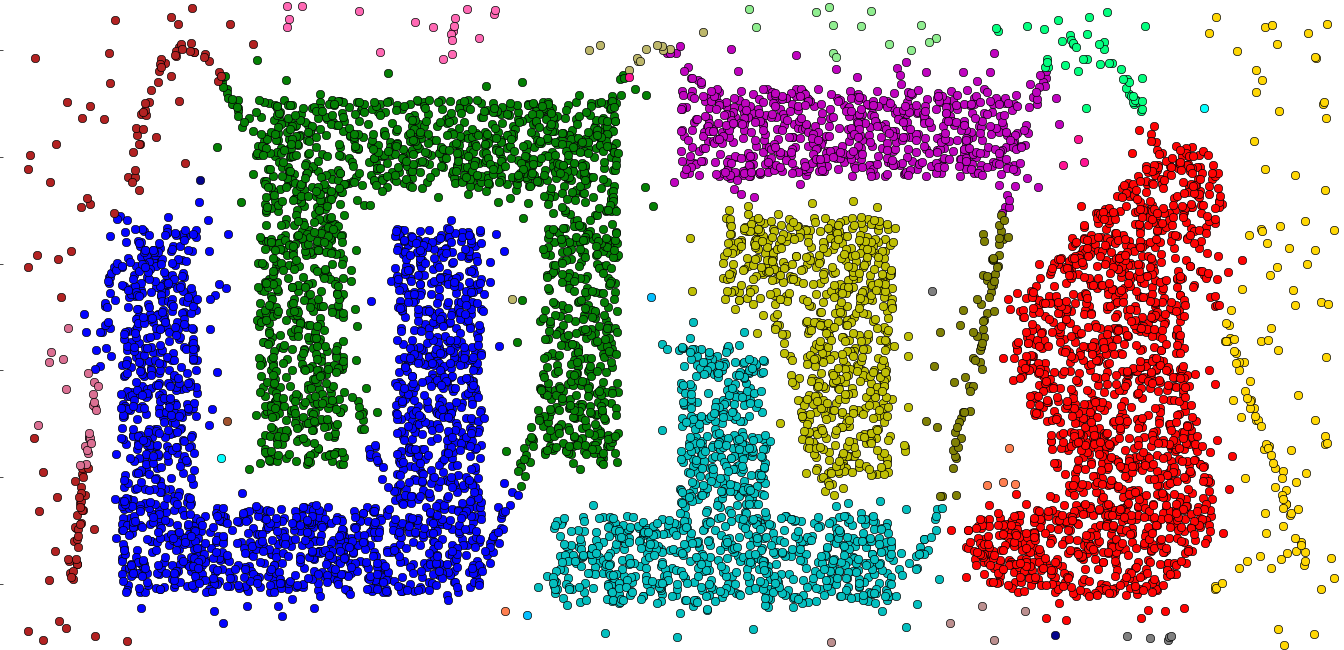}
  \caption{AWC result for \(DS4\) (\( n= 10000  \)) and \( DS3 \) (\( n= 8000 \)) with \( \lambda = 15 \).}
  \label{fig:chameleonawc}
\end{figure}

%


\begin{figure}[t]
  \includegraphics[width=\linewidth]{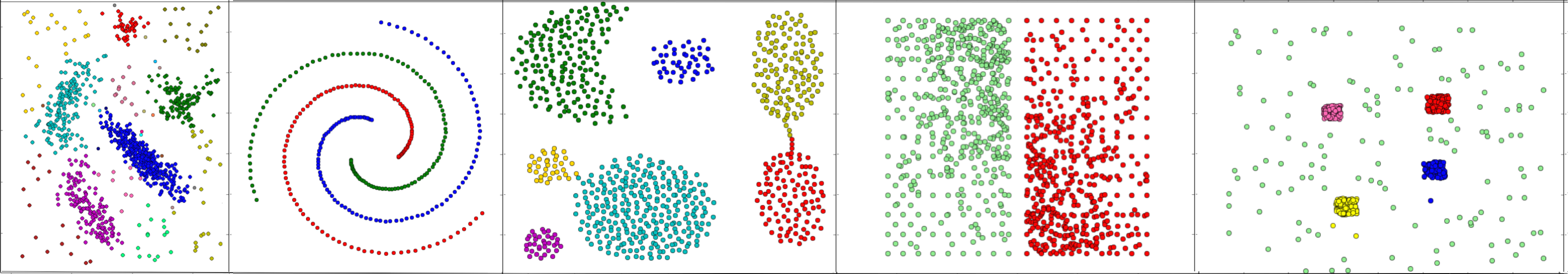}
  \caption{AWC result for artificial datasets}
  \label{fig:manyart}
\end{figure}

\subsection{Text data}

This section demonstrates the performance of AWC on text
 data, where the data dimension is very large. 
In our experiments we used 9 text datasets from the CLUTO toolkit \cite{karypis2002cluto} which are widely used in the literature.
 The basic characteristics of the datasets are summarized in Table \ref{table:restext}.
The datasets dimension \( \dimd \) ranges from 2886 to 10128 which makes these datasets a good benchmark for testing AWC manifold property on high-dimensional data. 
CLUTO provides already preprocessed datasets. This preprocessing includes stop-word removal and stemming.
In our experiments we represent the documents using the traditional vector space model with TF-IDF transformation: \(i\)-th document is presented as vector \( X_i = \{ x_{ij} \}_{j=1}^d \)  where
\begin{EQA}[c]
	x_{ij} = tf_{ij} \times idf_{j}, 
	\qquad
	idf_{j} \eqdef \log(1 + n) - \log (1 + n_{j}) + 1.
\end{EQA}
Here \( tf_{ij} \) is the frequency of term \( j \) in the document \( i \),  \( n_{j} \) is the number of documents which contains the term \( j \) and \( idf_{j} \) is the inverse document frequency.
The last one reflects how important a word is to a document in a collection.
%
%
Originally the six datasets \textit{tr11, tr12, tr23, tr31, tr41}, and \textit{tr45} are derived from TREC collections 
(Text Retrieval Conference, http://trec.nist.gov).
The datasets \textit{re0} and \textit{re1} are taken from Reuters-21578 text categorization test collection \cite{lewis1997reuters}.
The dataset \textit{wap} is from the WebACE project \cite{boley1999document}
where each document corresponds to a web page listed in the subject hierarchy of Yahoo!. 
%
For evaluation we used Normalized Mutual Information NMI from \cite{strehl2002cluster},
which is a popular measure for clustering accuracy in text data literature.
For a true clustering structure \( \CL^{*} = \{\CL^{*}_{m}\}_{m=1}^M \) and some other structure 
\( \CL = \{\CL_{l}\}_{l=1}^{L}\), define
\( n_{ml} = |\CL^{*}_{m} \cap \CL_{l}| \), \( n_{m}^{*} = |\CL^{*}_{m}| \),
\( n_{l} = |\CL_{l}| \) and
\begin{EQA}[c]
	\NMI(\CL, \CL^{*})
	=
	\frac{\sum_{ml} n_{ml} \log{\frac{n \, n_{ml}}{n_{m}^{*} \, n_{l}}}}
		 {\sqrt{\sum_{m} n_{m}^{*} \log( n_{m}^{*}/n) \, \, \sum_{l} n_{l} \log(n_{l}/n)}} \, .
\end{EQA} 
%
We compared AWC with state-of-the-art algorithms for clustering textual data:
Spectral clustering with Normalized Cut (NCut)
 \cite{shi2000normalized};
 Local Learning based Clustering Algorithm (LLCA) \cite{wu2006local};
 Clustering via Local Regression (CLOR), \cite{sun2008clustering};
 Regularized Local Reconstruction for Clustering (RLRC) \cite{sun2009regularized}.
 These methods
belong to the group of spectral clustering approaches. 
The results for these algorithms on our benchmark are taken from the work \cite{sun2009regularized}. 
All methods were provided with correct number of clusters \( K \) and the neighborhood size \( k = 40 \). 
For more details about experimental settings we refer to \cite{sun2009regularized}. 
The LLCA results are obtained after tuning the parameters.
We also use the procedure CLUTO which is a bisecting graph partitioning-based algorithm. 
The package 
 \textit{vcluster} from CLUTO toolkit \cite{karypis2002cluto} was used 
with the prespecified correct number of clusters \( K \) and a neighborhood size \( k = 40 \).
 Other parameters were set by default. 
The results of all methods are presented in Table \ref{table:restext}. 
The maximal two numbers in each row are marked in bold.
For CLUTO, after 100 runs, we calculated the best and the worst result among these runs, they are presented in the corresponding column of Table \ref{table:restext} in the form \([\NMI_{worst}, \NMI_{best}]\).

AWC also have a starting neighborhood size \( n_0 = 40 \) which is similar to other methods. 
The Euclidean distance was used as similarity measure.
Another parameter of AWC is the effective dimension
\( \dime \) used in \eqref{uniltdef} for computing the values \( \unil_{ij}^{(k)} \).
We just set \( \dime=2\). 
In the Table \ref{table:restext} the result of AWC after tuning  \( \lambda \) is marked by \( \AWC^{*} \).
By \( \AWC_{s} \) we marked the result obtained with \( \lambda \) chosen by ``sum-of-weights'' heuristic. One can see that in most cases ``sum-of-weights'' heuristic result is close to optimal. 
Table \ref{table:restext} shows that in 
7 out of 9 datasets (\textit{tr11, tr23, tr31, tr45, re0, re1, wap}) the result of AWC is similar 
to the best result among all considered state-of-the-art algorithms. 
It is worth mentioning that 
all methods except AWC were provided with the correct number of clusters \( K \).
In addition all considered algorithms were constructed specially for text data whereas AWC is remained unchanged and all results in this and other sections are obtained by the same algorithm.

\begin{table}
\caption{\label{table:restext} NMI for real-world text data sets, two best results in each row are in bold.}
\begin{tabular}{llll llll lll}
&
\multicolumn{7}{c}{Algorithm} & & &\\
\cline{2-8}
Data &  \( \AWC^{*} \) & \( \AWC_{s} \) & 
NCut
&LLCA
&CLOR
&RLRC
&CLUTO
& \(n\) & \(d\) & \( K \)
 \\ \hline
\multirow{1}{*}{\textit{tr11}}
 & \textbf{0.715}   & 0.712  &  
 0.624 & 0.620 & 0.674 & \textbf{0.723}
 &  [0.637, 0.706 ]  
         &  414 & 6429  &  9 \\ \hline
 \multirow{1}{*}{\textit{tr12}} 
 &  0.652 & 0.624 &  
 0.599 & 0.622 & 0.657 & \textbf{0.737}
 &  [0.632, \textbf{0.732} ]  
         &  313 & 5804  &  8 \\ \hline
  \multirow{1}{*}{\textit{tr23}} 
 &  \textbf{0.457}  & 0.34  &  
0.344 & 0.298 & 0.334 & 0.357 
&   [0.409, \textbf{0.446} ] 
         &  204 &  5832 &  6 \\ \hline
  \multirow{1}{*}{\textit{tr31}} 
 & \textbf{0.641 }  & {0.602} &  
0.457 & 0.499 & 0.483 & 0.534 
&  [0.615, \textbf{0.661} ]  
         &  927 &  10128 &  7\\ \hline
  \multirow{1}{*}{\textit{tr41}} 
 &  0.639  & 0.585 &  
 0.603 & 0.622 & \textbf{0.642} & 0.620 
 &   [0.639, \textbf{0.697} ] 
         &  878 &  7454 &  10  \\ \hline
  \multirow{1}{*}{\textit{tr45}} 
 &  \textbf{0.72}  & {0.682}  &  
 0.558 & 0.585 & 0.631 & 0.664   
 &  [0.605, \textbf{0.708} ]  
         &  690 &  8261 &  10 \\ \hline
  \multirow{1}{*}{\textit{re0}} 
 &  \textbf{0.468}  & \textbf{0.458}  &  
0.401 & 0.409 & 0.426 & 0.395   
&  [0.367, 0.429 ]  
        & 1504  & 2886  & 12 \\ \hline
  \multirow{1}{*}{\textit{re1}} 
 &  \textbf{0.609}  & {0.583} &  
0.484 & 0.485 & 0.498 & 0.496  
&  [0.555, \textbf{0.607} ]  
         &  1657 & 3758  & 24 \\ \hline
  \multirow{1}{*}{\textit{wap}} 
 & \textbf{0.598}  &  {0.586}  &  
 0.525 & 0.542 & 0.541 & 0.577  
 &  [0.578, \textbf{0.611} ]  
         &  1560 & 8460  & 19 \\ \hline
\end{tabular}
\qquad \qquad \qquad
\qquad \qquad \qquad
\qquad \qquad \qquad
\qquad \qquad \qquad
%
\end{table}




\def\uquotq{q}
\def\uquotQ{\rho}
\def\lagr{\mathcal{L}}
\def\ccS{\cc{S}}

\section{Proofs}
\label{SAWCproofs}
This section presents the proofs of the main results.
First we show that the value \( \tilde{\theta}_{ij}^{(k)} \) 
is a root-n consistent estimator of the gap coefficient \( \theta_{ij}^{(k)} \) 
for any two neighbor balls \( \B(X_{i},h_{k}) \) and \( \B(X_{j},h_{k}) \). 
Unlike standard results from empirical process theory, this bound 
is dimension free and does not involve any entropy number. 
The proof mainly uses combinatorial arguments.

\begin{lemma}
\label{LEPTkij}
For any \( k \leq K \) and any \( i \ne j \) with \( \dist(X_{i},X_{j}) \leq h_{k} \), let 
the gap coefficient \( \theta_{ij}^{(k)} \) be defined by \eqref{thetaijkBXiBXj} 
and its estimate \( \tilde{\theta}_{ij}^{(k)} \) by \eqref{thetaijkdefNN}. 
Then it holds for 
a fixed constant \( \zz \) on a random set of probability
at least \( 1 - 2 \ex^{-\zz} \)
\begin{EQA}
	\P\left( N_{\iju}^{(k)} \,\, \kullb(\tilde{\theta}_{ij}^{(k)},\theta_{ij}^{(k)}) 
		> \zz
	\right)
	& \leq &
	2 \ex^{-\zz} .
\label{NijoKttijktijk2ez}
\end{EQA}
\end{lemma}

\begin{proof}
Let us fix a step \( k \) and a pair of points \( X_{i}, X_{j} \) with 
\( \dist(X_{i},X_{j}) \leq h_{k} \).
Without loss of generality, we assume \( i=1 \) and \( j=2 \).
Denote 
\begin{EQA}[rclcrcl]
	\B_{12} 
	& \eqdef &
	\B(X_{1},h_{k}) \cup \B(X_{2},h_{k}) ,
	&\qquad&
	\cc{O}_{12} 
	& \eqdef &
	\B(X_{1},h_{k}) \cap \B(X_{2},h_{k}) .
\label{B12kO12kdef}
\end{EQA}
Given \( X_{1},X_{2} \) the remaining observations \( X_{3},\ldots,X_{n} \) are
still i.i.d. from the same distribution. 
Let also \( \ccS \) be the index subset of the set \( \bigl\{ 3,\ldots,n \bigr\} \).
Introduce the random event \( A_{\ccS} \) by conditions 
\( X_{\ell} \in \B_{12} \)
for \( \ell \in \ccS \) and \( X_{\ell} \not\in \B_{12} \) for 
\( \ell \in \ccS^{c} \eqdef \bigl\{ 3,\ldots,n \bigr\} \setminus \ccS \):
\begin{EQA}
	A_{\ccS}
	& \eqdef &
	\bigl\{ X_{\ell} \in \B_{12}, \ell \in \ccS,
		X_{\ell} \not\in \B_{12}, \ell \in \ccS^{c}
	\bigr\} .
\label{ASkXlB12SSc}
\end{EQA}
After conditioning on \( X_{1}, X_{2} \) and on \( A_{\ccS} \),
the subsample \( \bigl\{ X_{\ell} \bigr\}_{\ell \in \ccS} \) 
is still i.i.d. with the conditional density 
\( \dens(x)/\P\bigl( A_{\ccS} \cond X_{1},X_{2} \bigr) \).
Therefore, the \( \xi_{\ell} = \Ind\bigl( X_{\ell} \in \cc{O}_{12} \bigr) \)'s
are given \( X_{1}, X_{2}, A_{\ccS} \) i.i.d. Bernoulli with the parameter 
\( \theta_{\ccS} = \theta_{12}^{(k)} \).
The deviation bound from \cite{PoSp2006} implies for the normalized sum 
\( \tilde{\theta}_{\ccS} \eqdef  N_{\ccS}^{-1} \sum_{\ccS} \xi_{\ell} \)
with \( N_{\ccS} \eqdef |\ccS| \):
\begin{EQA}
	\P\Bigl( 
		N_{\ccS} \kullb\bigl( \tilde{\theta}_{\ccS}, \theta_{\ccS} \bigr) > \zz 
		\cond X_{1},X_{2}, A_{\ccS} 
	\Bigr)
	& \leq &
	2 \ex^{-\zz} ;
	\quad
	\zz \geq 0.
\label{PNSKttStS2ez}
\end{EQA}
As the right hand-side of this inequality does not depend on \( X_{1},X_{2} \),
\( \ccS \), and \( A_{\ccS} \), the bound applies for the joint distribution
in the unconditional form yielding \eqref{NijoKttijktijk2ez}.
\end{proof}

\paragraph{Proof of Theorem~\ref{Tpropconcdens}}
\label{SProofpropde}
Suppose that the density function \( \dens(x) \) fulfills one of two theorem conditions.
Let also all the weights \( \weight_{ij}^{(m)} \) for \( m < k \) 
computed at the first \( k-1 \) steps of the algorithm are equal to one.
It remains to show that the next step \( k \) leads to the same results. 
Our inductive assumption means that we consider non-adaptive weights 
\( \weight_{ij}^{(k)} \) which only account to the distance between 
points \( X_{i} \), \( X_{j} \), and \( X_{\ell} \) for all \( \ell \ne i,j \)
with \( \dist(X_{i},X_{\ell}) \leq h_{k} \) or 
\( \dist(X_{j},X_{\ell}) \leq h_{k} \).
Now Lemma~\ref{LEPTkij} ensures \eqref{NijoKttijktijk2ez} for any pair \( X_{i}, X_{j} \) with \( \dist(X_{i},X_{j}) \leq h_{k} \) and any \( k \geq 1 \). 
Also by Lemma~\ref{Ltijkqijk}, it holds \( \theta_{ij}^{(k)} \geq \unil_{ij}^{(k)} \). 
For the event \( \tilde{\theta}_{ij}^{(k)} < \unil_{ij}^{(k)} \leq \theta_{ij}^{(k)} \) we are interested in,
this implies by convexity of the Kullback-Leibler divergence w.r.t. the first argument that 
\begin{EQA}
	\P\Bigl( N_{\iju}^{(k)} \, \kullb \bigl(\tilde{\theta}_{ij}^{(k)},\unil_{ij}^{(k)} \bigr) 
		\Ind\bigl( \tilde{\theta}_{ij}^{(k)} < \unil_{ij}^{(k)} \bigr) > \zz	
	\Bigr) 
 	& \leq &
	\P\Bigl( N_{\iju}^{(k)} \, \kullb \bigl(\tilde{\theta}_{ij}^{(k)},\theta_{ij}^{(k)} \bigr) > \zz \Bigr) 
	\leq 
	2 \ex^{- \zz} .
\label{PNijukKLttsijk}
\end{EQA}
This implies a uniform bound: for an absolute constant \( \CONST \leq 4 \)
\begin{EQA}
	\P \left( \max_{i \ne j} \max_{k \geq 1}
		N_{\iju}^{(k)} \, \kullb \bigl(\tilde{\theta}_{ij}^{(k)},\unil_{ij}^{(k)} \bigr) 
		\Ind\bigl( \tilde{\theta}_{ij}^{(k)} < \unil_{ij}^{(k)} \bigr) > \CONST \log n  
	\right)
	& \leq &
	\frac{2}{n} \, .
\label{PmaxijkNijk1n}
\end{EQA}

\paragraph{Proof of Theorem~\ref{TmainCLseplb}}
Let \( V \) be a set with the volume \( |V| \) and \( G \) be a splitting hole 
with volume \( |G| \)
such that \( \VsG \eqdef V \setminus G \) consists of two disjoint regions.
Let also the data density be equal to \( p_{G} \) on \( G \) and to \( p_{\VsG} \) 
on the complement \( V \setminus G \).
Consider two hypothesis \( H_{0} \) of ``no gap'' \( p_{G} = p_{\VsG} = 1/|V| \) 
and \( H_{G} \) of a \( G \)-gap with \( p_{G} = (1 - \eps) p_{\VsG} \).
We are interested to understand the conditions which enable us to separate these two hypotheses.
Define \( \delta = |G|/|V| \), so that \( |\VsG| / |V| = 1 - \delta \).
Then under \( H_{0} \) the data distribution is uniform on the set \( V \) with the density
\( p_{0} = 1/|V| \).
Further, under \( H_{G} \) the data density is uniform on \( G \) 
with the density \( p_{G} \) 
and on its complement \( V \setminus G \) 
with the density \( p_{\VsG} \) satisfying 
\begin{EQA}
	|\VsG| p_{\VsG} + |G| p_{G} 
	&=& 
	(1 - \delta) |V| p_{\VsG} + \delta |V| (1-\eps) p_{\VsG}
	=
	1, 
\label{11VsGpVsGGpG}
\end{EQA}
yielding
\begin{EQA}
	p_{\VsG}
	=
	\frac{p_{0}}{1 - \delta \eps},
	& \quad &
	p_{G}
	=
	\frac{p_{0} (1 - \eps)}{1 - \delta \eps} .
\label{pVsGp01med}
\end{EQA}
For the experiment with \( N \) observations, 
the condition of consistent separation between \( H_{0} \) and \( H_{G} \) is that 
the total Kullback-Leibler (KL) divergence between two distributions converges to infinity. 
The KL divergence for the model with \( N \) i.i.d. observations is defined as 
\( \kullb(\P_{0},\P_{G}) = \E_{0} \log (d\P_{0}/d\P_{G}) \).
As \( \P_{0}(G) = |G| p_{0} = |G|/|V| = \delta \), it follows by \eqref{pVsGp01med}
\begin{EQA}
	\kullb(\P_{0},\P_{G})
	&=&
	N \P_{0}(G) \log\frac{p_{0}}{p_{G}} + N \P_{0}(\VsG) \log \frac{p_{0}}{p_{\VsG}}
	\\
	&=&
	N \delta \log\frac{1 - \delta \eps}{1-\eps} 
	+ N (1 - \delta) \log ( 1 - \delta \eps ) 
	=
	N \log ( 1 - \delta \eps ) - N \delta \log(1 - \eps).
\label{kullbCLNln2P0PG}
\end{EQA}
If \( G \) is a hole of a fixed volume \( \delta |V| \) and \( \eps = \eps_{N} \to 0 \), then  
\begin{EQA}
	\kullb(\P_{0},\P_{G})
	&=&
	0.5 (\delta - \delta^{2}) N \eps_{N}^{2} \bigl\{ 1 + O( \eps_{N}) \bigr\}
\end{EQA}
and consistent separation between \( P_{0} \) and \( P_{G} \) is impossible if 
\( N \eps_{N}^{2} \) remains bounded
by a fixed constant as \( N \) grows.

\paragraph{Proof of Theorem~\ref{TmainCLsep}}
Now we show that the AWC algorithm does a good job in detecting a gap between 
two neighbor clusters separated by a hole 
\( G \) of the volume \( |G| = \delta |V| \) and the piecewise constant density 
given by \eqref{pVsGp01med}.
Let \( V \) consist of three neighbor regions of equal cylindric shape of height \( h \)
and base radius \( \rho h \) for some \( \rho < 1 \).
The hole \( G \) corresponds to the central part, so that \( |G| = |V|/3 \)
and \( \delta = 1/3 \).
We consider two points \( X_{i}, X_{j} \) from different side of the hole separated by a distance 
\( \| X_{i} - X_{j} \| \geq h_{k} \geq h \) at the step \( k \). 
Due to the definition, it is sufficient to show that the corresponding test statistic 
\( \testst_{ij}^{(k)} \) exceeds \( \lambda \).
We sketch the proof of this fact for the ``worst case'' situation that the procedure 
did not gain any structural information during the first \( k-1 \) steps and
all the earlier computed adaptive weights \( \weight_{il}^{(k-1)} \) and 
\( \weight_{jl}^{(k-1)} \) coincide with the non-adaptive distance based weights,
i.e. they 
are equal to one within the balls of radius \( h_{k-1} \) around these points.
By the theorem conditions, it holds
\begin{EQ}[rcl]
	A_{\iju}
	\eqdef
	\B(X_{i},h_{k}) \cup \B(X_{j},h_{k}) 
	& = &
	V,
	\quad
	A_{\ijo} \supset G ,
\label{AijuuGAijusG}
\end{EQ}
and the value \( \unil_{ij}^{(k)} = |A_{\ijo}| / |V| \) satisfies
\( 1/3 \leq \unil_{ij}^{(k)} \leq 1/2 \).
As \( \P(A_{\iju}) = \P(V) = 1 \), it holds
\begin{EQA}
	\theta_{ij}^{(k)}
	=
	\P(A_{\ijo})
	&=&
	p_{G} |G| + p_{\VsG} |A_{\ijo} \setminus G| \, ,
	\\
	&=&
	\bigl\{ |G| (1 - \eps) + (|A_{\ijo}| - |G|) \bigr\}	p_{\VsG}
	=
	\frac{ |A_{\ijo}| - \eps |G| }{|V| (1 - \delta \eps)}
	=
	\frac{q_{ij}^{(k)} - \delta \eps}{1 - \delta \eps} \, 
\label{PGAijoqijde}
\end{EQA}
yielding
\begin{EQA}
	\unil_{ij}^{(k)} - \theta_{ij}^{(k)}
	&=&
	\frac{(1 - \unil_{ij}^{(k)}) \delta \eps}{1 - \delta \eps} 
	\geq  
	\CONST \eps \,
\label{qijPAijoeps}
\end{EQA}
with \( \CONST \geq 1/6 \).
In particular, this means that \( \theta_{ij}^{(k)} < \unil_{ij}^{(k)} \).
Also one can bound 
\begin{EQA}
	\kullb^{1/2}\bigl( {\theta}_{ij}^{(k)},\unil_{ij}^{(k)} \bigr)
	& \geq &
	\CONST_{1} \eps
\label{kulltijkqijkCe2}
\end{EQA}
with a slightly different constant \( \CONST_{1} \).
To show that \( \tilde{\theta}_{ij}^{(k)} \) is significantly smaller than 
\( \unil_{ij}^{(k)} \), we apply Lemma~\ref{LEPTkij}.
The condition \( A_{\iju} = V \) implies \( N_{\iju} = n \) and by 
Lemma~\ref{LEPTkij} 
\begin{EQ}[rcl]
	n \, \kullb\bigl( \tilde{\theta}_{ij}^{(k)},\theta_{ij}^{(k)} \bigr)
	& \leq &
	\CONST_{2} \log(n) \, .
\label{NijoNNijcNxi}
\end{EQ}
If \( \tilde{\theta}_{ij}^{(k)} \leq {\theta}_{ij}^{(k)} \), then
\( \kullb\bigl( \tilde{\theta}_{ij}^{(k)},\unil_{ij}^{(k)} \bigr) \geq 
\kullb\bigl( {\theta}_{ij}^{(k)},\unil_{ij}^{(k)} \bigr) \).
If \( {\theta}_{ij}^{(k)} < \tilde{\theta}_{ij}^{(k)} \leq \unil_{ij}^{(k)} \), then
regularity and convexity of \( \kullb(x,\unil) \) w.r.t. \( x,\unil \) implies
\begin{EQA}
	\kullb^{1/2}\bigl( \tilde{\theta}_{ij}^{(k)},\unil_{ij}^{(k)} \bigr)
	& \geq &
	\fis \, \kullb^{1/2}\bigl( {\theta}_{ij}^{(k)},\unil_{ij}^{(k)} \bigr)
	- \kullb^{1/2}\bigl( \tilde{\theta}_{ij}^{(k)},\theta_{ij}^{(k)} \bigr) 
\label{KL12ttijkqijk}
\end{EQA}
for some fixed constant \( \fis > 0 \);
see \cite{PoSp2006} for more details.
This together with \eqref{kulltijkqijkCe2} and \eqref{NijoNNijcNxi} implies 
\begin{EQA}
	\kullb^{1/2}(\tilde{\theta}_{ij}^{(k)},\unil_{ij}^{(k)}) 
	& \geq &
	\fis \CONST_{1} \eps - \sqrt{{\CONST_{2} n^{-1} \, \log n} } 
	\geq 
	\sqrt{{\lambda}/{n}}
\label{kullbttijkqijkCe2}
\end{EQA}
provided that \( \fis \CONST_{1} \eps \sqrt{n} \geq \sqrt{\CONST_{2} \log(n)} + \sqrt{\lambda} \).
Together with the bound \( \lambda \leq \CONST \log(n) \) this yields 
a consistent separation \( \weight_{ij}^{(k)} = 1 \) under condition 
\( \eps^{2} \geq \CONST n^{-1} \log(n) \).

\section{Conclusion}

The proposed procedure AWC systematically exploits the idea of extracting the structural information
about the underlying data distribution from the observed data in terms of adaptive weights 
and uses this information for sensitive clustering.
The method is appealing and computationally feasible, 
the numerical results indicate the state-of-the-art performance of the method. 
Theoretical results show its optimality in separating of neighbor regions.



\bibliography{clusterbib,listpubm-with-url}


\clearpage
\pagebreak
\begin{appendices}
\appendix
\section{Technical proofs}
\label{SAppndA}
\begin{lemma}
\label{lemma:testgap}
Let \( X_{1},\ldots,X_{n} \in \R^{\dimp} \) be an i.i.d. sample and \( B, C \) be two 
non-overlapping measurable sets in \( \R^{\dimp} \).
For a given value \( \uquotq \in (0,1) \), define one sided hypotheses
\begin{EQA}
	H_{0} 
	&:&
	\P(B) \geq \uquotq (\P(B) + \P(C)),
	\\
	H_{1}
	&:& 
	\P(B) < \uquotq (\P(B) + \P(C)).
\end{EQA}
Then the likelihood-ratio test statistic \( T \) for testing 
the null hypothesis \( H_0 \) against the  alternative \( H_{1} \) is given by
\begin{EQA}
	T 
	&=&
	(S_{B} + S_{C}) \, \kullb\bigl( \tilde{\theta},\unil \bigr) \,
	\bigl\{ \Ind( \tilde{\theta} \leq \unil) - \Ind( \tilde{\theta} > \unil) \bigr\},
\end{EQA}
where \( \kullb(\theta,\eta) \) is the Kullback-Leibler (KL) divergence between 
two Bernoulli laws with parameters \( \theta \) and \( \eta \):
\begin{EQA}
	\kullb(\theta,\eta)
	& \eqdef &
	\theta \log \frac{\theta}{\eta} + (1 - \theta) \log \frac{1 - \theta}{1 - \eta} \, 
\label{kullbthetaetadefu}
\end{EQA}
and  
\begin{EQA}
	\tilde{\theta}
	&=&
	\frac{S_{B}}{S_{B} + S_{C}} \, .
\label{titetSBSBSC}
\end{EQA}
\end{lemma}
\begin{proof}

Define \( A \) as the complement of \( B \) and \( C \): \( A \eqdef (B \cup C)^{c} \). 
Let also \( a = \P(A) \), \( b = \P(B) \), \( c = \P(C) \), 
and 
\begin{EQA}
	S_{A}
	&\eqdef & 
	\sumi \Ind(X_{i} \in A),
	\quad
	S_{B}
	\eqdef
	\sumi \Ind(X_{i} \in B),
	\quad
	S_{C}
	\eqdef 
	\sumi \Ind(X_{i} \in C) .
\label{SAIndXiA}
\end{EQA}
The log-likelihood \( L(a,b,c) \) for the multinomial model with the parameter \( (a,b,c) \) reads
\begin{EQA}
	L(a,b,c)
	&=&
	S_{A} \log a + S_{B} \log b + S_{C} \log c + R ,
\label{LabcSASBSC}
\end{EQA}
where the remainder \( R \) does not depend on \( a,b,c \) from \( [0,1] \).
Now for a fixed \( \uquotQ \in [0,1] \), define
\begin{EQA}
	\hat{L}(\uquotQ)
	& \eqdef &
	\sup_{a+b+c=1, \, b = \uquotQ (b + c)} L(a,b,c) .
\label{hLrhoabc1Labc}
\end{EQA}
Then, the maximum likelihood under null hypothesis \( H_0 \) can be written as
\begin{EQA}
	\hat{L}_{0}
	& \eqdef &
	\sup_{1 > \uquotQ \geq \uquotq} \hat{L}(\uquotQ).
\end{EQA}
Under the alternative
\begin{EQA}
	\hat{L}_{1}
	& \eqdef &
	\sup_{0 < \uquotQ < \uquotq} \hat{L}(\uquotQ), 
\end{EQA}
and the likelihood ratio test statistic \( T \)
is defined as the difference between \( \hat{L}_{1} \) and \( \hat{L}_{0} \):
\begin{EQA}
	T
	&\eqdef &
	\hat{L}_{1} - \hat{L}_{0} \, .
\end{EQA}
Introduce also the quantity 
\begin{EQA}
	\hat{L}
	& = &
	\sup_{a+b+c = 1} L(a,b,c)
\label{hKdefrhoma}
\end{EQA}
Optimization under the constraint \( a+b+c = 1 \) yields
in view of \( S_{A} + S_{B} + S_{C} = n \) that
\begin{EQA}
	\hat{L}
	&=&
	S_{A} \log \frac{S_{A}}{n}
	+ S_{B} \log \frac{S_{B}}{n}
	+ S_{C} \log \frac{S_{C}}{n} + R \, .
\label{hatLabcSAlogSAn}
\end{EQA}
It is also easy to see that
\begin{EQA}
	\hat{L}
	& = &
	\max_{\uquotQ}\hat{L}(\uquotQ)
	=
	\hat{L}(\tilde{\theta})
\end{EQA}
with \( \tilde{\theta} \) from \eqref{titetSBSBSC}.

Similar optimization under the additional constraint 
\( b = \uquotQ (b + c) \) (see below at the end of the proof) 
\begin{EQA}
	\hat{L}(\uquotQ)
	& \eqdef &
	\sup_{a+b+c=1, \, b = \uquotQ (b + c)} L(a,b,c)
	\\
	&=&
	S_{A} \log \frac{S_{A}}{n} 
	+ (S_{B} + S_{C}) \log \frac{S_{B} + S_{C}}{n}
	+ S_{B} \log \uquotQ + S_{C} \log (1-\uquotQ) \, .
\label{optineqABCabc}
\end{EQA}

Consider the derivative of \( \hat{L}(\uquotQ) \):
\begin{EQA}
	\frac{\partial \hat{L}(\uquotQ)}{\partial \uquotQ}
	&=&
	\frac{S_{B} }{\uquotQ} - \frac{S_{C} }{1-\uquotQ}
	=
	\frac{S_{B} - (S_{B} + S_{C}) \uquotQ}{\uquotQ(1-\uquotQ)}
	=
	\frac{(S_{B} + S_{C})}{\uquotQ(1-\uquotQ)} (\tilde{\theta} - \uquotQ).
\label{titetSBSBSC}
\end{EQA}
It follows
\begin{EQA}
	\frac{\partial \hat{L}(\uquotQ)}{\partial \uquotQ}
	> 0
	& \iff &
	0 < \uquotQ < \tilde{\theta}
	\\
	\frac{\partial \hat{L}(\uquotQ)}{\partial \uquotQ}
	< 0
	& \iff &
	1 > \uquotQ > \tilde{\theta} \, .
\end{EQA}
To calculate \( \hat{L}_{0}, \hat{L}_{1} \) we need to consider two cases:
\begin{EQA}
	\uquotq \leq \tilde{\theta}
	& \implies &
	\hat{L}_{0} = \hat{L}, \quad 
	\hat{L}_{1} = \hat{L}(\uquotq)
	\\
	\uquotq > \tilde{\theta}
	& \implies &
	\hat{L}_{0} = \hat{L}(\uquotq), \quad 
	\hat{L}_{1} = \hat{L} .
\end{EQA}
The likelihood ratio test statistic is defined as the difference between \( \hat{L} \)
and \( \hat{L}_{0} \):
\begin{EQA}
	T
	&\eqdef &
	\hat{L}_{1} - \hat{L}_{0}
	=
	\bigl\{ \hat{L} - \hat{L}(\uquotq) \bigr\} \,
	\bigl\{ \Ind( \tilde{\theta} \leq \unil) - \Ind( \tilde{\theta} > \unil) \bigr\}
	\\
	&=&
	(S_{B} + S_{C}) \Bigl\{ \tilde{\theta} \log \frac{\tilde{\theta}}{\unil} 
		+ (1 - \tilde{\theta}) \log \frac{1 - \tilde{\theta}}{1 - \unil} 
	\Bigr\} \,
	\bigl\{ \Ind( \tilde{\theta} \leq \unil) - \Ind( \tilde{\theta} > \unil) \bigr\}.
\label{TesthatLhatL0SASB}
\end{EQA}
Note that this test statistic can be written as
\begin{EQA}
	T 
	&=&
	(S_{B} + S_{C}) \kullb\bigl( \tilde{\theta},\unil \bigr)
	\, \bigl\{ \Ind( \tilde{\theta} \leq \unil) - \Ind( \tilde{\theta} > \unil) \bigr\}
\label{TSBSCkullbttu}
\end{EQA}
as required.

It remains to check \eqref{optineqABCabc}. 
The Lagrange function  for this optimization problem reads as follows
\begin{EQA}
	\lagr(a,b,c, \nu, \mu)
	&=&
	S_{A} \log a + S_{B} \log b + S_{C} \log c 
	-
	\nu (a + b + c - 1)
	-
	\mu (b - \uquotQ (b + c)) .
\end{EQA}
The partial derivatives of the Lagrange function are:
\begin{EQA}[c]
\begin{dcases}
	\frac{\partial \lagr}{\partial a}
	=
	\frac{S_{A}}{a} - \nu = 0
	\\
	\frac{\partial \lagr}{\partial b}
	=
	\frac{S_{B}}{b} - \nu - \mu (1 - \uquotQ) = 0
	\\
	\frac{\partial \lagr}{\partial c}
	=
	\frac{S_{C}}{c} - \nu + \mu \uquotQ = 0
	\\
	\frac{\partial \lagr}{\partial \nu}
	=
	a + b + c - 1 = 0
	\\
	\frac{\partial \lagr}{\partial \mu}
	=
	b - \uquotQ (b + c) = 0 .
\end{dcases}
\end{EQA}
These equations can be rewritten as follows:
\begin{EQA}[c]
	\begin{dcases}
		a 
		=
		\frac{S_{A}}{\nu}
		\\
		b
		=
		\frac{S_{B}}{\nu + \mu (1 - \uquotQ)}
		\\
		c
		=
		\frac{S_{C}}{\nu - \mu \uquotQ}
		\\
		1 = a + b + c 
		\\
		c = \frac{b(1-\uquotQ)}{\uquotQ}
	\end{dcases}
\end{EQA}
Combining second, third and fifth equations 
\begin{EQA}
	\frac{S_{C}}{\nu - \mu \uquotQ}
	&=&
	\frac{(1-\uquotQ)}{\uquotQ}
	\frac{S_{B}}{\nu + \mu (1 - \uquotQ)}
	\\
	\mu
	&=&
	-\nu 
	\frac{\uquotQ (S_{B} + S_{C}) - S_{B}}{\uquotQ(1-\uquotQ)(S_{B} + S_{C})}
\end{EQA}
and
\begin{EQA}
	b
	&=&
	\nu^{-1} 
	\frac{S_{B}}{1 - \frac{\uquotQ (S_{B} + S_{C}) - S_{B}}{\uquotQ(S_{B} + S_{C})}}
	=
	\frac{\uquotQ(S_{B} + S_{C})}{\nu}
	\\
	c
	&=&
	\nu^{-1} 
	\frac{S_{C}}{1 + \frac{\uquotQ (S_{B} + S_{C}) - S_{B}}
	{(1-\uquotQ)(S_{B} + S_{C})}}
	=
	\frac{(1-\uquotQ)(S_{B} + S_{C})}{\nu} \, .
\end{EQA}
It follows from \( a+b+c=1 \):
\begin{EQA}[c]
	\frac{S_{A}}{\nu}
	+
	\frac{\uquotQ(S_{B} + S_{C})}{\nu}
	+
	\frac{(1-\uquotQ)(S_{B} + S_{C})}{\nu}
	=
	1
	\\
	\nu = S_{A} + S_{B} + S_{C} = n .
\end{EQA}
Finally we derive
\begin{EQA}
	a 
	&=&
	\frac{S_{A}}{n}
	\\
	b
	&=&
	\frac{\uquotQ(S_{B} + S_{C})}{n}
	\\
	c
	&=&
	\frac{(1-\uquotQ)(S_{B} + S_{C})}{n}
\end{EQA}
which yields the assertion.
\end{proof}

Our next lemma helps to check that in a region with a linear or univariate concave 
density, the gap coefficient for any two overlapping balls 
is not smaller than the one corresponding to the uniform density. 
Here we assume that \( \dist(X_{i},X_{j}) = \| X_{i} - X_{j} \| \).
\begin{lemma}
\label{Ltijkqijk}
Consider the situation with a linear or univariate concave density \( \dens(x) \) 
for \( x \in V \).
For any two points \( X_{i},X_{j} \in V \) with \( \dist( X_{i} , X_{j}) \leq h_{k} \),
the gap coefficient \( \theta_{ij}^{(k)} \) from \eqref{thetaijkBXiBXj} fulfills
\begin{EQA}
	\theta_{ij}^{(k)}
	& \geq &
	\unil_{ij}^{(k)},
\label{unildensxxch}
\end{EQA}
where the value \( \unil_{ij}^{(k)} \) corresponds to a constant density 
\( \dens_{0}(x) \equiv \CONST \). 
\end{lemma}

\begin{proof}
We write \( h \) in place of \( h_{k} \) for ease of notation.
In the case of a linear density, 
it holds \( \theta_{ij}^{(k)} = \unil_{ij}^{(k)} \) by symmetricity arguments.
In the case of a univariate concave density 
%
consider a linear function \( g(x) \) such that it
 coincides with \( \dens(x) \) in points
  \( X_{i}-h  \) and   \( -X_{i}+h  \):  
  \( g(X_{i}-h) = \dens(X_{i}-h),
  g(-X_{i}+h) = \dens(-X_{i}+h) \).
   Concavity of \( \dens(x) \) implies
   \begin{EQA}
	\dens(x) \geq g(x), &\quad& x \in A \eqdef [X_{i}-h,-X_{i}+h] ,
	\\
	\dens(x) \leq  g(x), &\quad& x \in B  \eqdef [-X_{i}-h, X_{i}+h] \setminus [X_{i}-h,-X_{i}+h] .
\end{EQA}
It follows
\begin{EQA}
	\frac{\int_{A} \dens(x) dx}{\int_{A} \dens(x) dx + \int_{B} \dens(x) dx}
	& \geq & 
	\frac{\int_{A} g(x) dx}{\int_{A} g(x) dx + \int_{B} \dens(x) dx}
	\geq
	\frac{\int_{A} g(x) dx}{\int_{A} g(x) dx + \int_{B} g(x) dx}
	=
	q .
\end{EQA}
This yields the result.
\end{proof}

One more result specifies the case of manifold structure for the underlying density.

\begin{lemma}
\label{lem:manprop}
Fix a \( d_\Pi \)-dimensional hyper-plane \( \Pi \in \R^{\dimp} \), \( \dimp_{\Pi} < \dimp \).
Consider the manifold \( M \) in \( \R^{\dimp} \) which can be represented as 
\begin{EQA}
M
&=&
\bigcup_{x \in \Omega} \Pi_{x}
\end{EQA}
where \( \Omega \in \R^{\dimp} \) is a convex set of dimension 
\( \dimp_{\Omega} \leq \dimp - \dimp_{\Pi}  \) and diameter \( h \), such that subspace 
of \( \Omega \) is
orthogonal to \( \Pi \).
\( \Pi_x \) is a shifted hyper-plane  \( \Pi \) such that \( x \in \Pi_{x} \).
Consider two points \( O_{1}, O_{2} \in M \) and two balls
\( \B_{1} = \B(O_{1}, R)\), \( \B_{2} = \B(O_{2}, R) \) with radius \( R \) 
and centers in \( O_{1}, O_{2} \). 
Then for \( R \gg h \) it holds
\begin{EQA}[c]
\frac{V_d((\B_{1} \cap \B_{2}) \cap M)}
{V_d((\B_{1} \cup \B_{2}) \cap M)}
\approx
q_{\dimp_{\Pi}} > q_{\dimp}
\end{EQA}
where \( q_{\dimp} \) is equal to \( q\left(\frac{|O_{1}O_{2}|}{R} \right)\) 
from \eqref{uniltdef} with the corresponding 
dimension \( \dimp \), \(|O_{1}O_{2}|\) is the distance between \( O_{1}, O_{2} \),
\( V_{\dimp} \) is \( \dimp \)-dimensional volume.
\end{lemma}
\begin{proof}
The considered case is represented on Figure \ref{fig:sep5}. 
Then 
\begin{EQA}
	\frac{V_d((\B_{1} \cap \B_{2}) \cap M)}{V_{\dimp}((\B_{1} \cup \B_{2}) \cap M)}
	&=&
	\frac{\int_{x \in \Omega} V_{\dimp_{\Pi}}((\B_{1} \cap \B_{2}) \cap \Pi_{x}) \, dx}
		 {\int_{x \in \Omega} V_{\dimp_{\Pi}}((\B_{1} \cup \B_{2}) \cap \Pi_{x})}
	\approx ({R \gg h})
	\\
	&\approx &
	\frac{V_{\dimp_{\Omega}}(\Omega) V_{\dimp_{\Pi}}((\B_{1} \cap \B_{2}) \cap \Pi_{O_{1}})}
		 {V_{\dimp_{\Omega}}(\Omega) V_{\dimp_{\Pi}}((\B_{1} \cup \B_{2}) \cap \Pi_{O_{1}})}
	=
	\frac{V_{\dimp_{\Pi}}(\B_{1}^{\dimp_{\Pi}} \cap \B_{2}^{\dimp_{\Pi}}) }
		 {V_{\dimp_{\Pi}}(\B_{1}^{\dimp_{\Pi}} \cap \B_{2}^{\dimp_{\Pi}}) }
	=
	q_{\dimp_{\Pi}} \, ,
\end{EQA}
where \( \B_{1}^{\dimp_{\Pi}}, \B_{2}^{\dimp_{\Pi}} \) are the balls in \( \R^{\dimp_{\Pi}}\)
with radii \( R \) and distance between centers \( |O_{1}O_{2}| \).
From equation \eqref{uniltdef} it follows: 
\( d_{\Pi} < \dimp \Rightarrow q_{\dimp_{\Pi}} > q_{\dimp} \).
\end{proof}

\section{Fixing the sequence \( h_{k} \)}
\label{SAppndhk}
A sequence \( h_{k} \) ensuring
\begin{EQA}[c]
	n(X_{i},h_{k+1})
	\leq 
	\hrate \, n(X_{i},h_{k}) ,
	\,
	h_{k+1} \leq b \, h_{k} ,
	\label{condition}
\end{EQA}
 can be fixed as follows. 
Let us collect for each point \( X_{i} \) the distances \( h_{\ell}(X_{i}) \) between \( X_{i} \) and its \( n_{\ell} \)-s neighbor, \( \ell =1, \dots, M \). 
In the homogeneous case, all \( h_{\ell}(X_{i}) \) for a fixed \( \ell \) and different \( i \) 
are of the same order. 
However, one can often observe a high variability of such radii in the inhomogeneous situation. 
Let a set \( \bigl\{ h_{\ell}^{*} \, , \ell \geq 0 \bigr\} \) be obtained by puling all series 
\( \bigl\{ h_{\ell}(X_{i}), \, \ell = 0,1,\ldots,\kmax \bigr\} \) together and by putting 
them in the increasing order. 
We will select the radii \( h_{k} \) sequentially from this set to ensure the condition \eqref{condition}.
Set \( h_{0} = h_{0}^{*} \). 
Equivalently, \( h_{0} \) is the smallest radius among all \( h_{0}(X_{i}) \). 
Then select the largest index \( \ell_{1} \) such that 
\begin{EQA}
	\max_{i} \frac{n(X_{i},h_{\ell_{1}}^{*})}{n(X_{i},h_{0})}
	& \leq &
	\hrate 
\label{maxinXihlish2}
\end{EQA}
and set \( h_{1} = h_{\ell_{1}}^{*} \). 
The construction of sequences \( \{ h_{\ell} \} \) ensures that such \( \ell_{1} > 1 \) exists. 
Continue in this way. If \( h_{k} = h_{\ell_{k}} \) is the radius selected at step \( k \), then
the next radius \( h_{k+1} \) is selected using the largest index \( \ell_{k+1} > \ell_{k} \) such that 
\( h_{k+1} = h_{\ell_{k+1}}^{*} \) ensures the condition. 
Stop when \( h_{k} \) reaches the largest possible value \( h_{\kmax} \). 
The condition  \eqref{condition} can be weaken by just controlling the fraction of points 
for which the inequality \eqref{condition} can be violated.

\section{Other clustering procedures}
\label{SAppndC}
Here we briefly describe the details how the concurring procedures were implemented. 
For algorithms with a univariate tuning parameter we took 100 evenly spaced values from
the prespecified range. 
Finally, the best result of each algorithm is used for comparison.

For \( k \)-\emph{means clustering} the best result is chosen from
100 runs of algorithm for each \( k: \)\( 1 \leq k \leq 3K\), where \( K \) is the true number of clusters taken from the data.


\emph{DBSCAN}  \cite{ester1996density} takes \textit{eps} and \textit{minsp} as the parameter combination to determine dense points, where \textit{eps} is the maximum distance between two samples to be considered in the same neighborhood, and \textit{minsp} is minimum number of points required to form a dense region.
For the best result of DBSCAN we evaluated over \( \textit{eps} \in [mindist, maxdist]\) and  \( \textit{minsp} \in [1, N]\), where \textit{maxidst}(\textit{mindist}) is the maximum(minimum) pairwise distance between the data elements and \( N \) is the data set size. 
DBSCAN can identify points as noise which are colored black on figures. The noise is considered as separate cluster.

\emph{Spectral clustering} constructs affinity matrix using either kernel function such the Gaussian (RBF) kernel or a k-nearest neighbors connectivity matrix \cite{zelnik2004self}. 
For the first case, the scaling factor  \( \sigma\) and  \textit{degree} of RBF kernel are tuned by varying over  \( \sigma \in [mindist, maxdist]\) and \textit{degree} up to 4.
For the second case, the parameter for number of neighbors \(  n \in [1, N]\) is tuned. For each parameter value, the best result from 100 runs with random initialization is taken.
As a final result the best output of all cases is taken.

\emph{Affinity propagation} has two parameters for tuning: dumping factor \( D \) from [0.5, 1] and preferences \( P \) for each point to be chosen as exemplars \cite{frey2007clustering}.  
We set \( P \) to a global shared value varying from minimum to maximum value of pairwise similarities (negative Euclidean distance) between data points. 
The adjustment of these parameters was rather difficult because of high sensitivity 
of the results to the parameter choice.
%
%

\section{Examples on separation ability of AWC}
\label{Ssepwithgapnum}
\label{SAppndSG}
Here we consider the case of two dense clusters \( A \) and \( C \) separated by an area of lower density \( B \). 
Explicitly we consider a rectangle with sizes \( 2 \times 3 \) and three area inside it presented on the Figure \ref{fig:gapexp}. 
The left and right areas have the same density \( p \), while the area in between has density 
\( \dens_{\eps} = (1-\eps) \dens\), \( \eps \in [0,1] \). 
An example of such generated data with \( \eps=0.3 \), \( n=1000 \) is shown on the middle plot of the Figure \ref{fig:gapexp}; in the last plot true clusters are labeled by colors. 
\begin{figure}[h]
\centering
  \includegraphics[width=0.19\textwidth]{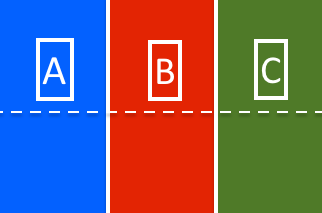}
  \quad
  \includegraphics[width=0.19\textwidth]{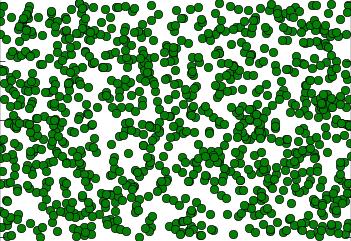}
  \quad
  \includegraphics[width=0.19\textwidth]{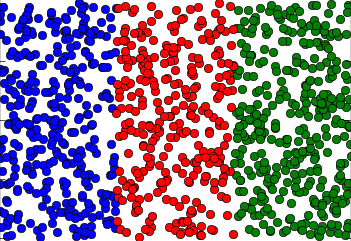}
  \caption{From left: clusters' areas, realization, true clustering. \( \eps = 0.3, n=1000 \)}
  \label{fig:gapexp}
\end{figure}

We expect that AWC separates the left and right clusters. 
In this experiment we are not interested in the behavior of AWC in between.
To measure how good AWC separates the clusters A and C we will use the separation error \( e_{s} \):
\begin{EQA}[c]
e_{s} 
	=
	\frac{\sum\limits_{i \neq j} |\hat{\weight}_{ij}| \Ind_{( \weight^{*}_{ij} = 0 )} \Ind_{(i,j \in A \cup C)}}
		 {\sum\limits_{i \neq j} \Ind_{( \weight^{*}_{ij} = 0 )}\Ind_{(i,j \in A \cup C)}},
\end{EQA}
where \( \weight^{*}_{ij} \) are true weights and \( \hat{\weight}_{ij} \) are answer weights of AWC.

Let us fix the overall number of points \( n \) and the parameter \( \eps  \). 
After running 200 experiments we can calculate the average separation error \( e_s (n, \eps) \).
For all experiments we count which part of them has error \( e_{s} > 0.1 \). 
For each \(n\) the probability having separation error \( e_{s} > 0.1 \) as a function of  \( \eps \) is shown on the right plot of Figure  \ref{fig:gapmainplot}.
On the left plot of Figure \ref{fig:gapmainplot} we show for each number of points \(n\) what difference in density \( \eps \) we can detect such that it guaranties probability of error level
 \( e_{s} > 0.1 \) being less than 0.1. E.g. for \(n = 1000\) the value \( \eps = 0.47 \) guaranties that the probability of   \( e_{s} > 0.1 \) is less than 0.1.
For each \( n \) the parameter \( \lambda \) was chosen to have average propagation error \( e_p\) equal 0.1. Hereby run the procedure on data with \( n \) points and \( \eps = 0 \) and take the minimum \( \lambda \) with propagation error \( e_p\) = 0.1. True clustering in this case is one cluster containing all points.

\begin{figure}[t]
\centering
\setlength{\abovecaptionskip}{-3pt}
  \includegraphics[width=0.45\textwidth]{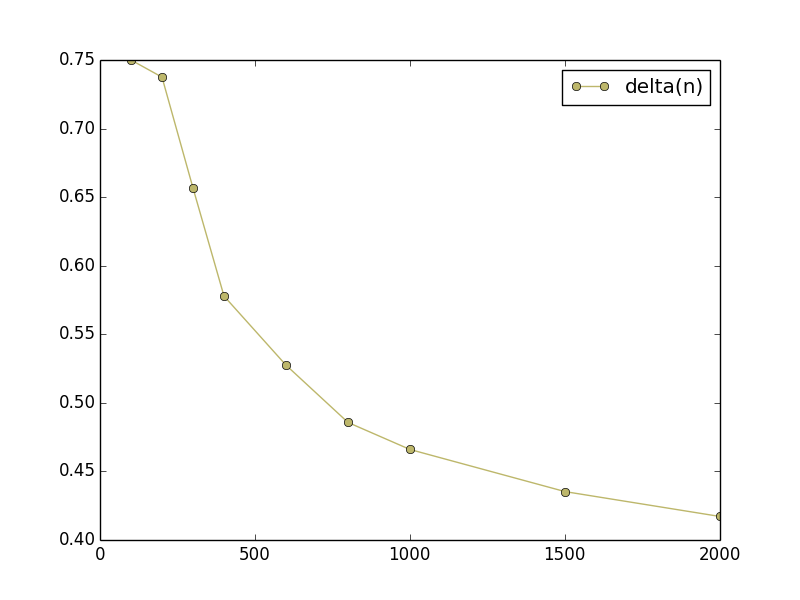}
  \qquad
  \includegraphics[width=0.45\textwidth]{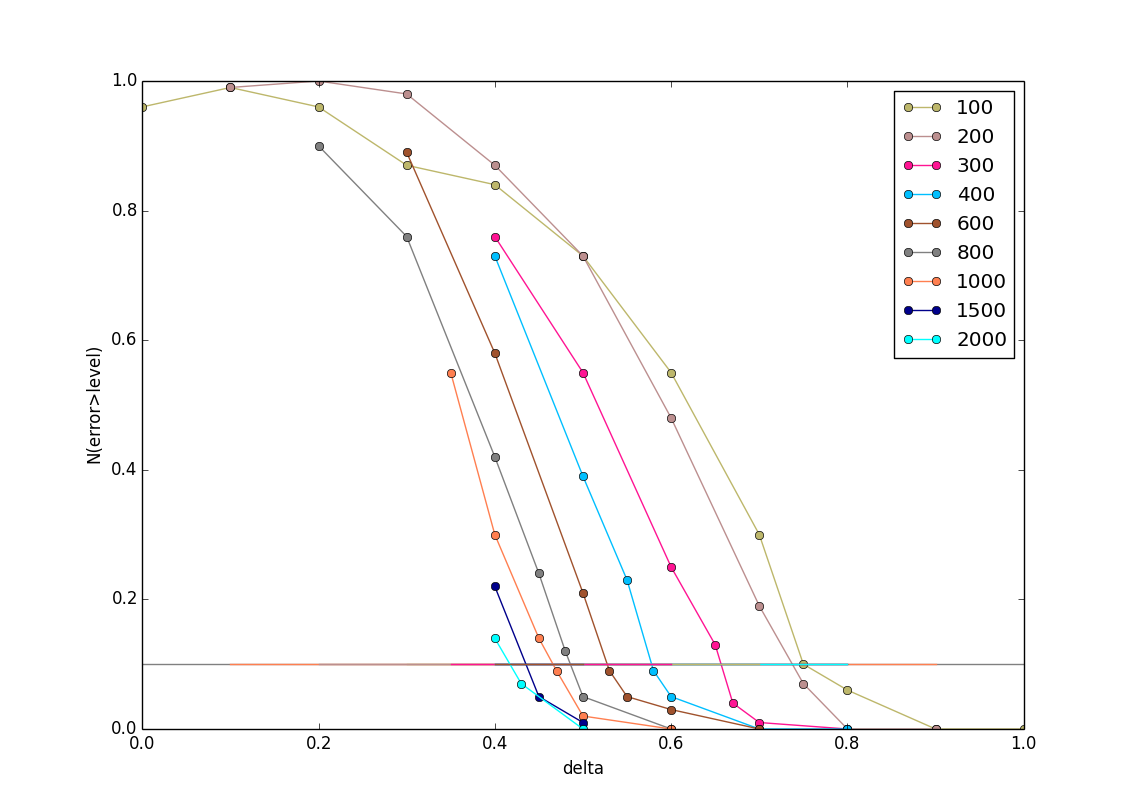}
  \caption{The smallest density gap \( \eps(n) \) yielding \( \P(e_s >0.1) \leq 0.1 \) 
  for different \( n \)
  and 
  \( \P(e_s >0.1) \) for different \( n \) and the gap coefficient \( \eps \).}
  \label{fig:gapmainplot}
\end{figure}

\section{Experiments on Real World datasets}
\label{SAppndRW}

The quality of the method depends on its separation and propagation ability. 
The separation quality of the method can be regarded as its ability to separate
different clusters. 
The propagation quality is considered as its ability to aggregate points from the same cluster. 
As the method is formulated in terms of weights, 
it is natural to measure these two types of misweighting error via the final computed weights
\( \hat{\weight}_{ij} \):  
\( e_{s} \) counts all connections (positive weights) between points from different clusters,
while \( e_{p} \) indicates the number of disconnecting points in the same cluster:
\begin{EQA}
	e_{s} 
	=
	\frac{\sum\limits_{i \neq j} |\hat{\weight}_{ij}| \Ind_{( \weight^{*}_{ij} = 0 )}}
		 {\sum\limits_{i \neq j} \Ind_{( \weight^{*}_{ij} = 0 )}},
	& \quad &
	e_{p} 
	=
	\frac{\sum\limits_{i \neq j} |1 - \hat{\weight}_{ij}| \Ind_{( \weight^{*}_{ij} = 1 )}}
		 {\sum\limits_{i \neq j} \Ind_{( \weight^{*}_{ij} = 1 )}},
	\quad
\label{errorsdef}
\end{EQA}
where \( \weight_{ij}^{*} \)  
denote the true weights describing the underlying clustering structure.
The \(e_{p}\) and \(e_{s}\) are just weighted parts of the well known metric 
for cluster analysis comparison called 
\textit{Rand index} \( R \) \cite{rand1971objective}. In our notation rand 
index can be represented
as
\begin{EQA}[l]
	R
	=
	1
	-
	\frac{\sum\limits_{i \neq j} |\hat{\weight}_{ij}| \Ind_{( \weight^{*}_{ij} = 0)}
			+ \sum\limits_{i \neq j} |1 - \hat{\weight}_{ij}| \Ind_{( \weight^{*}_{ij} = 1 )}
		}
		{\sum\limits_{i \neq j} \Ind_{( \weight^{*}_{ij} = 0 )} 
			+ \sum\limits_{i \neq j} \Ind_{( \weight^{*}_{ij} = 1 )}}
\eqdef
1 - e.\quad
\label{randindex}
\end{EQA}
Further we will use the \textit{general error} \( e \eqdef 1 - R \) instead of Rand index.

Now  consider the behavior of the algorithms on real world data.
The data sets are taken from UCI repository
\cite{Lichman:2013} , except the \textit{Olive} data; see \\
http://www2.chemie.uni-erlangen.de/publications/ANN-book/datasets/oliveoil/.
\textit{Iris} data set contains 3 type of iris plants.
\textit{Wine} data is the result of a chemical analysis of wines grown in the same region in Italy but derived from three different cultivars. 
\textit{Seeds} comprise kernels belonging to three different varieties of wheat.
\textit{Thyroid} is clinical data used to predict  patients thyroid functional state.
 \textit{Ecoli} data is used for classification of the cellular localization sites of proteins. 
\textit{Olive}  is a group of olive oil samples from nine different regions of Italy.
\textit{Wisconsin} stands for Wisconsin Breast Cancer Database designated whether samples are benign or malignant. 
\textit{Banknote} data set was extracted from images that were taken from genuine and forged banknote-like specimens.
 The data set sizes \( n \),  number of attributes \( d \) and  clusters
 \( K \) are listed in Table \ref{table:res}.


Similarly to the experiments on artificial data, 
each algorithm was run for best parameter configuration which minimizes the general error \( e \).
Algorithms performances are listed in Table \ref{table:res}. 
Here for every algorithm only general error \( e \) is presented.
The graphical interpretation of the Table \ref{table:res} is shown on Figure \ref{fig:comp}.  
Here x-axis represent the error level and y-axis shows the number of databases.
%
\begin{figure}
\centering
  \includegraphics[width=0.6\textwidth]{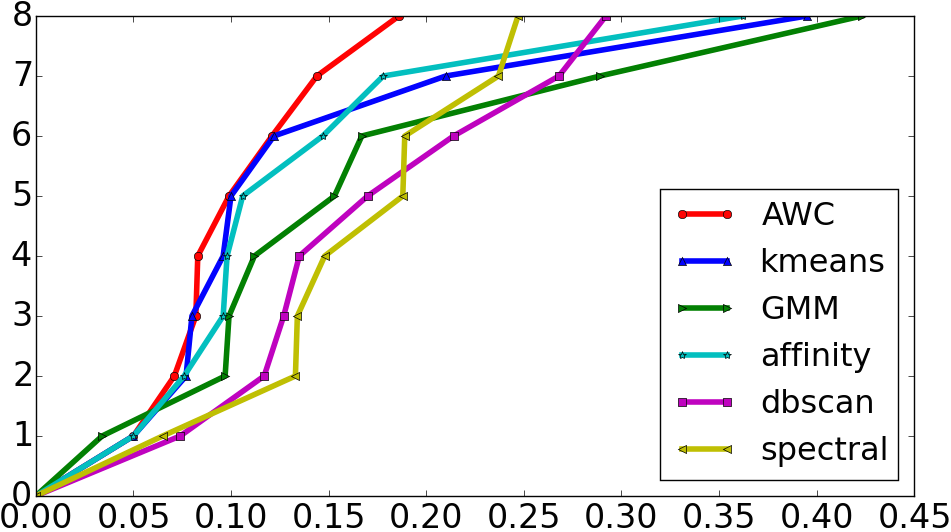}
  \caption{Comparison on real datasets}
  \label{fig:comp}
\end{figure}
For each clustering algorithm we construct its plot
as the function showing for any error threshold the number of databases where the
error level is below this threshold.
Thus each line is non-decreasing function and the best algorithm must lie on the left.
One can see that AWC demonstrates the best performance on the majority of databases.
The value of the sum of weights statistic \( S(\lambda) \) is shown on Figure
\ref{fig:platolive}.

\begin{figure}
\centering
  \includegraphics[width=0.49\textwidth]{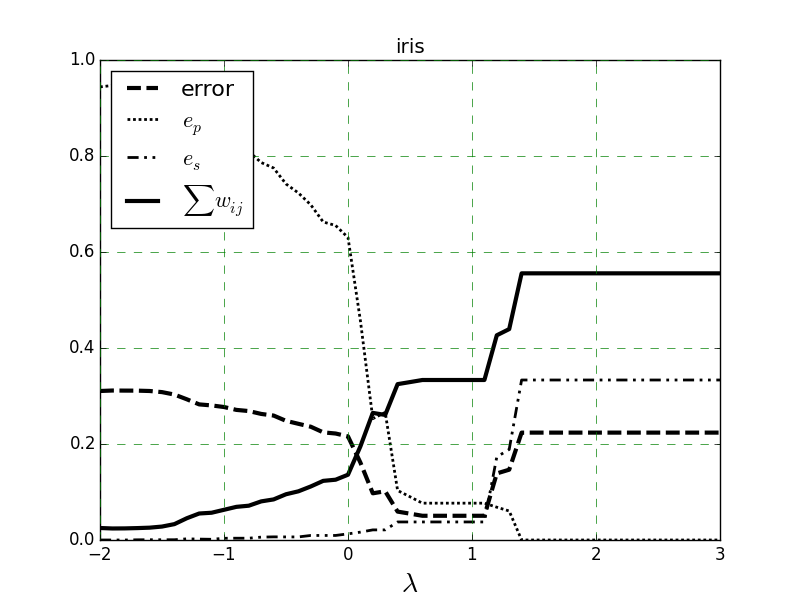}
  \includegraphics[width=0.49\textwidth]{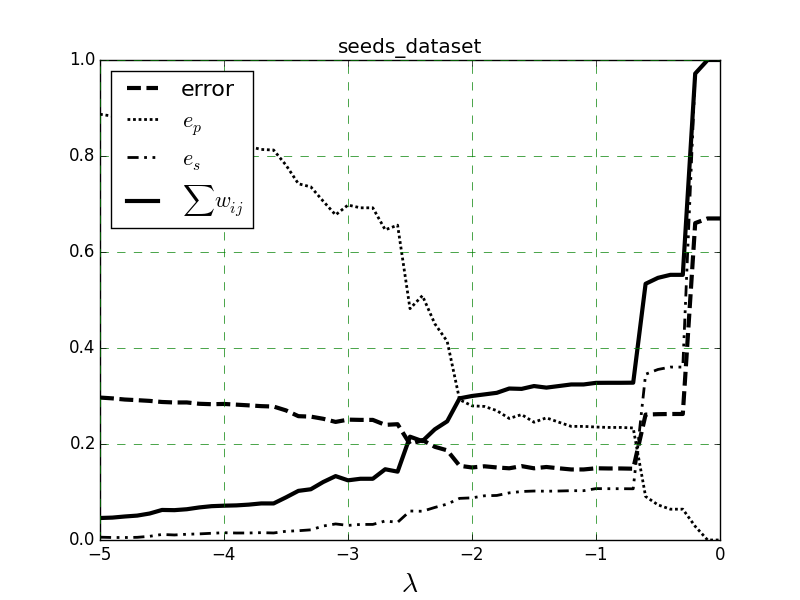}
\\
  \includegraphics[width=0.49\textwidth]{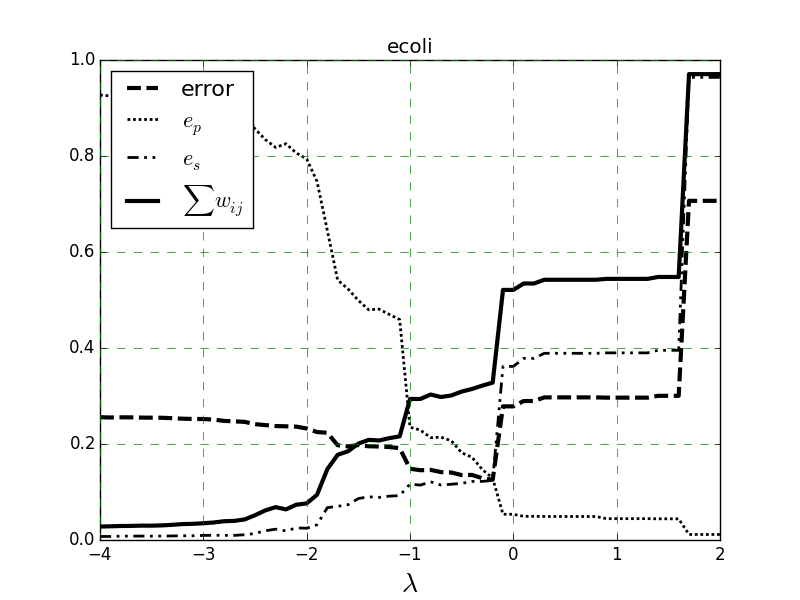}
  \includegraphics[width=0.49\textwidth]{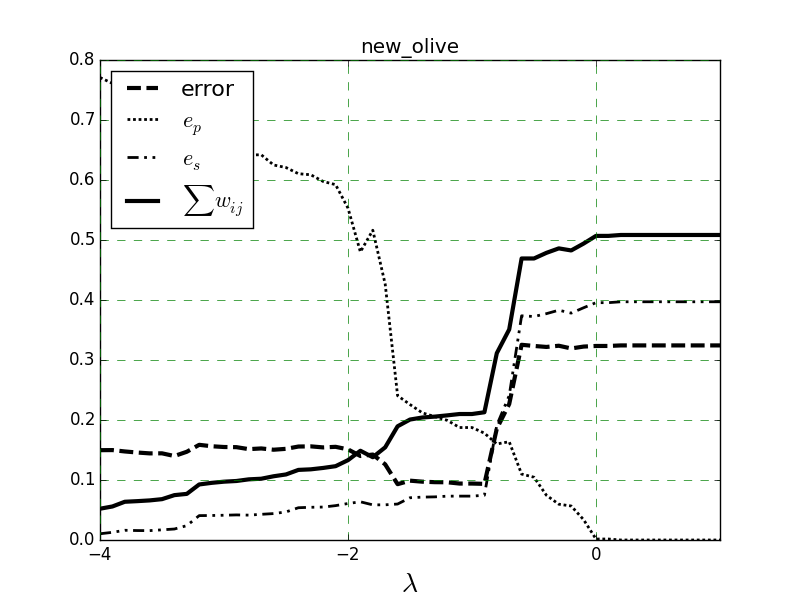}
\caption{\(S(\lambda)\) Iris (Top-link), Seeds (top-right), Ecoli (bottom-link), Olive(bottom-right)}
\label{fig:platolive}
\end{figure}

\begin{table}
\caption{\label{table:res} Real world data sets error \( e \) for each method,
the best two results are in bold}
\centering
\fbox{
\begin{tabular}{llll llll lll}
&
\multicolumn{7}{c}{Algorithm} & & & \\
\cline{2-8}
Data &  AWC & AWC\(_{sow}\) & k-m & GMM & Affinity & DBSCAN & Spectral 
& \(n\) & \(d\) & \( K \) \\ \hline
\multirow{1}{*}{Iris} 
 &   0.05 & 0.05 &  0.050 & \textbf{0.034} & 0.05 & 0.117 & 0.188 
& 150 & 4 & 3 \\ \hline
 \multirow{1}{*}{Wine} 
 &   \textbf{0.101} & 0.132 &  \textbf{0.096} & \textbf{0.099} & \textbf{0.096} & 0.268 & 0.189 
&178 & 13 & 3 \\ \hline
  \multirow{1}{*}{Seeds} 
 &   \textbf{0.148} & \textbf{0.148} &  0.21 & 0.289 & 0.178 & 0.292 & \textbf{0.148} 
 & 210 & 7 & 3 \\ \hline
 \multirow{1}{*}{Thyroid} 
 &   \textbf{0.089} &  \textbf{0.089} &  \textbf{0.08} & 0.097 & 0.147 & 0.135 & 0.247 
 & 215 & 5 & 3 \\ \hline
 \multirow{1}{*}{Ecoli} 
 &   0.125 &  0.125 &  0.122 & 0.167 & \textbf{0.106} & 0.17 & 0.134 
 & 336 & 7 & 8 \\ \hline
 \multirow{1}{*}{Olive} 
 &   0.093 & 0.093 &  0.1 & 0.153 & 0.076 & 0.127 & \textbf{0.065} 
 & 572 & 8 & 9 \\ \hline
 \multirow{1}{*}{Wisconsin} 
 &   \textbf{0.067} &  \textbf{0.070} &  \textbf{0.077} & 0.112 & 0.098 & \textbf{0.074} & 0.133 
 & 699 & 9 & 2 \\ \hline
  \multirow{1}{*}{Banknote} 
 &   \textbf{0.193}  & \textbf{0.194} &  0.395 & 0.423 & 0.362 & 0.214 & 0.237 
 & 1372 & 4 & 2 \\ \hline \hline 
\end{tabular}
}
\end{table}

\end{appendices}

\end{document}